\documentclass{article}

\usepackage{microtype}
\usepackage{graphicx}
\usepackage{subcaption}
\usepackage{booktabs} %
\usepackage{hyperref}
\usepackage{pbox}
\usepackage{bigints}
\usepackage{url}
\usepackage{comment}
\usepackage{booktabs}
\usepackage{enumitem}
\usepackage{wrapfig}
\usepackage{enumitem}
\usepackage{wrapfig}
\usepackage{tikz}
\usepackage{bbm, dsfont}
\mathchardef\mhyphen="2D
\usetikzlibrary{bayesnet}

\usepackage{color}
\usepackage[tableposition=top]{caption}
\usepackage[titletoc,toc,title]{appendix}
\usepackage[utf8]{inputenc} %
\usepackage[T1]{fontenc}    %
\usepackage{amsfonts}
\usepackage{amsthm}
\usepackage{nicefrac}
\usepackage{microtype}
\usepackage{graphicx}
\usepackage[raggedright]{sidecap}
\usepackage{xspace}
\usepackage{todonotes}
\usepackage{amsmath}
\usepackage{amsopn}
\usepackage{multirow}
\usepackage{multibib}
\newcites{sup}{References}

\usepackage{tikz}
\usetikzlibrary{bayesnet}

\newcommand{\FOMMpkpd}{\textbf{FOMM}_{\text{PK-PD}}}

\newcommand{\FOMMnl}{\textbf{FOMM}_{\text{NL}}}
\newcommand{\RNNpkpd}{\textbf{GRU}_{\text{PK-PD}}}
\newcommand{\RNN}{\textbf{GRU}}
\newcommand{\SSMpkpd}{\textbf{SSM}_{\text{PK-PD}}}

\newcommand{\SSMnl}{\textbf{SSM}_{\text{NL}}}
\newcommand{\FOMMmoe}{\textbf{FOMM}_{\text{MOE}}}
\newcommand{\SSMmoe}{\textbf{SSM}_{\text{MOE}}}

\newcommand{\R}[1]{\mathbb{R}^{#1}}

\newcommand{\FOMMlinear}{\textbf{FOMM}_{\text{Linear}}}

\newcommand{\SSMlinear}{\textbf{SSM}_{\text{Linear}}}

\def\IEF{\mu_{\theta}}
\def\PKPDIEF{\mathbb{PK\mhyphen PD}_{\text{Neural}}}

\theoremstyle{definition}

\theoremstyle{remark}

\PassOptionsToPackage{square,comma,numbers,sort&compress,super}{natbib}

\usepackage{xcolor}
\usepackage{xparse}

\ExplSyntaxOn

\keys_define:nn { miguel/label }
 {
  label   .tl_set:N = \l_miguel_label_tl,
  unknown .code:n   = \clist_put_right:Nx \l_miguel_label_clist
                       { \l_keys_key_tl = \exp_not:n { #1 } }
 }
\clist_new:N \l_miguel_label_clist
\box_new:N \l_miguel_label_box
\box_new:N \l_miguel_label_image_box

\NewDocumentCommand{\xincludegraphics}{O{}m}
 {
  \tl_clear:N \l_miguel_label_tl
  \clist_clear:N \l_miguel_label_clist
  \keys_set:nn { miguel/label } { #1 }
  \tl_if_empty:NTF \l_miguel_label_tl
   {
    \miguel_includegraphics:Vn \l_miguel_label_clist { #2 }
   }
   {
    \hbox_set:Nn \l_miguel_label_image_box
     {
      \miguel_includegraphics:Vn \l_miguel_label_clist { #2 }
     }
    \hbox_set:Nn \l_miguel_label_box
     {
      \skip_horizontal:n { 3pt }
      \fcolorbox{white}{white}{\footnotesize \tl_use:N \l_miguel_label_tl}
     }
    \leavevmode
    \box_use:N \l_miguel_label_image_box
    \skip_horizontal:n { -\box_wd:N \l_miguel_label_image_box }
    \hbox_overlap_right:n
     {
      \box_move_up:nn
       {
        \box_ht:N \l_miguel_label_image_box - 
        \box_ht:N \l_miguel_label_box - 3pt
       }
       { \box_use_drop:N \l_miguel_label_box }
     }
    \skip_horizontal:n { \box_wd:N \l_miguel_label_image_box }
   }
 }

\cs_new_protected:Nn \miguel_includegraphics:nn
 {
  \includegraphics[#1]{#2}
 }
\cs_generate_variant:Nn \miguel_includegraphics:nn { V }

\ExplSyntaxOff

\usepackage{comment}

\usepackage[accepted]{icml2021}

\icmltitlerunning{Neural Pharmacodynamic State Space Modeling}

\begin{document}

\twocolumn[
\icmltitle{Neural Pharmacodynamic State Space Modeling}

\icmlsetsymbol{equal}{*}

\begin{icmlauthorlist}
\icmlauthor{Zeshan Hussain}{equal,mit}
\icmlauthor{Rahul G. Krishnan}{equal,msr}
\icmlauthor{David Sontag}{mit}
\end{icmlauthorlist}

\icmlaffiliation{mit}{Massachussetts Institute of Technology, CSAIL and IMES, Cambridge, MA}
\icmlaffiliation{msr}{Microsoft Research New England, Cambridge, MA}

\icmlcorrespondingauthor{Zeshan Hussain}{zeshanmh@mit.edu}
\icmlcorrespondingauthor{Rahul G. Krishnan}{rahulgk@mit.edu}

\icmlkeywords{Machine Learning, ICML}

\vskip 0.3in
]

\printAffiliationsAndNotice{\icmlEqualContribution} %

\begin{abstract}
Modeling the time-series of high-dimensional, longitudinal data is important for predicting patient disease progression. However, existing neural network based approaches that learn representations of patient state, while very flexible, are susceptible to overfitting. We propose a deep generative model that makes use of a novel attention-based neural architecture inspired by the physics of how treatments affect disease state. The result is a scalable and accurate model of high-dimensional patient biomarkers as they vary over time. Our proposed model yields significant improvements in generalization and, on real-world clinical data, provides interpretable insights into the dynamics of cancer progression.
\end{abstract}

\section{Introduction\label{sec:introduction}}

Clinical biomarkers capture snapshots of a patient's evolving disease state as well as their response to treatment. However, these data can be high-dimensional, exhibit missingness, and display complex nonlinear behaviour over time as a function of time-varying interventions. Good unsupervised models of such data are key to discovering new clinical insights. This task is commonly referred to as disease progression modeling \citep{wang2014unsupervised,venuto2016review,schulam2016integrative,elibol2016cross,liu2015efficient,alaa2019attentive,severson2020}.

Reliable unsupervised models of time-varying clinical data find several uses in healthcare. One use case is enabling practitioners to ask and answer counterfactuals using observational data \citep{rubin1974estimating, pearl2009causal, bica2020estimating}. 
Other use cases include guiding early treatment decisions based on a patient's biomarker trajectory, detecting drug effects in clinical trials \cite{mould2007using}, and clustering patterns in biomarkers that correlate with disease sub-type \citep{zhang2019data}. 
To do these tasks well, understanding how a patient's biomarkers evolve over time given a prescribed treatment regimen is vital, since a person's biomarker profile is often the only observed proxy to their true disease state. Like prior work \citep{alaa2019attentive,severson2020,krishnan2017structured}, we frame this problem as a conditional density estimation task, where our goal is to model the density of complex multivariate time-series conditional on time-varying treatments. %

Representation learning exposes a variety of techniques for good conditional density estimation \cite{che2018recurrent,miotto2016deep,choi2016doctor,suresh2017clinical}.
For sequential data, a popular approach has been to leverage black-box, sequential models (e.g. Recurrent Neural Networks (RNNs)), where a time-varying representation is used to predict clinical biomarkers. Such models are prone to overfitting, particularly on smaller clinical datasets. More importantly, such models often make simplistic assumptions on how time-varying treatments affect downstream clinical biomarkers; for example, one choice is to concatenate treatments to the model's hidden representations \citep{alaa2019attentive,krishnan2017structured}. The assumption here is that the neural network learns how treatments influence the representation. We argue that this choice is a missed opportunity and better choices exist. Concretely, 
we aim to encourage neural models to learn representations that encode a patient's underlying disease burden by specifying how these representations evolve due to treatment.
We develop a new disease progression model that captures such insights by using inductive biases rooted in the biological mechanisms of treatment effect. 
Inductive biases have been integral to the success of deep learning in other domains such as vision, text and audio. 
For example, convolutional neural networks explicitly learn representations invariant to translation or rotation of image data \citep{lecun2012learning,jaderberg2015spatial,veeling2018rotation},
transformers leverage attention modules \citep{bahdanau2014neural,vaswani2017attention}
that mimic how human vision pays attention to various aspects of an image, and modified graph neural networks can explicitly incorporate laws of physics to generalize better \citep{seo2019differentiable}.
For some learning problems, the physics underlying the domain are often known, e.g. the laws of motion, and may be leveraged in the design of inductive biases \citep{ling2016reynolds,anderson2019cormorant,wang2020incorporating}. The same does not hold true in healthcare, since exact disease and treatment response mechanisms are not known. However, physicians often have multiple hypotheses of how the disease behaves during treatment. To capture this intuition, we develop inductive biases that allow for a data-driven selection over multiple neural mechanistic models that dictate how treatments affect representations over time.

\begin{figure}[t!]
\centering
\includegraphics[width=0.25\textwidth]{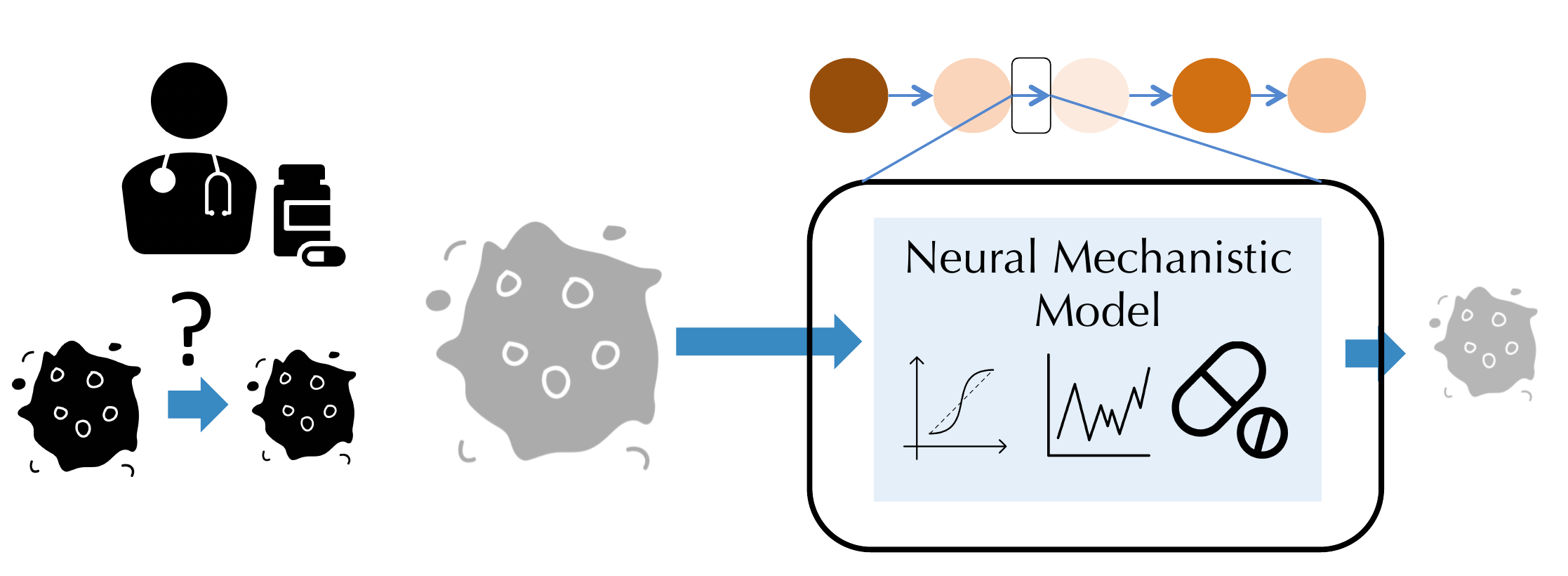}
\includegraphics[width=0.3\textwidth]{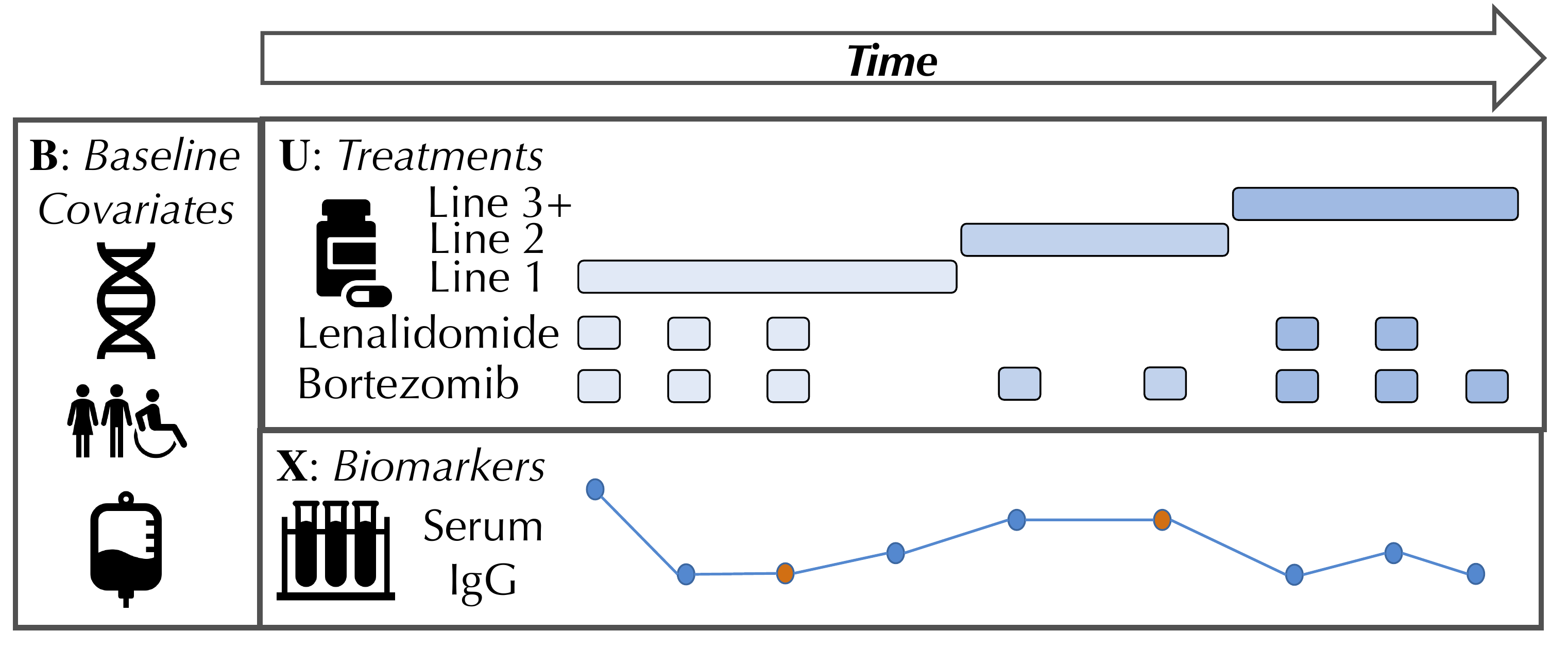}
\caption{\small
\textbf{Inductive Bias Concept (Top):} A clinician often has multiple mechanistic hypotheses as to how the latent tumor burden evolves. Our approach formalizes these hypotheses as neural architectures that specify how representations respond to treatments. 
\textbf{Patient Data (Bottom):} 
Illustration of data from a chronic disease patient. Baseline (static) data typically consists of genomics, demographics, and initial labs. Longitudinal data typically includes laboratory values (e.g. serum IgG) and treatments (e.g. lenalidomide). Baseline data is usually complete, but longitudinal measurements are frequently missing at various time points. 
}
\label{fig:data_vis}
\end{figure}

\textbf{Contributions:} We present a new attention-based neural architecture, $\PKPDIEF$, that captures the effect of drug combinations in representation
space (Figure \ref{fig:data_vis} [top]). It learns to attend over multiple competing mechanistic explanations of how a patient's genetics, past treatment history, and prior disease state influence the representation to predict the next outcome.
The architecture is instantiated in a state space model, $\SSMpkpd$, and shows strong improvements in generalization compared to several baselines and prior state of the art. We demonstrate the model can provide insights into multiple myeloma progression. Finally, we release a disease progression benchmark dataset called ML-MMRF, comprising a curated, pre-processed subset of data from the Multiple Myeloma Research Foundation CoMMpass study \citep{usrelating}. Our model code can be found at \url{https://github.com/clinicalml/ief}, and the data processing code can be found at \url{https://github.com/clinicalml/ml_mmrf}.

\section{Related Work}
\label{sec:related}

Much work has been done across machine learning, pharmacology, statistics and biomedical informatics on building models to characterize the progression of chronic diseases. Gaussian Processes (GPs) have been used to model patient biomarkers over time and estimate counterfactuals over a single intervention \citep{futoma2016predicting, schulam2017reliable, silva2016observational, soleimani2017treatment}. In each of these cases, the focus is either on a single intervention per time point or on continuous-valued interventions given continuously, both strong assumptions for chronic diseases. To adjust for biases that exist in longitudinal data, \citet{lim2018forecasting, bica2020estimating} use propensity weighting to adjust for time-dependent confounders. 
However, they concatenate multi-variate treatments to patient biomarkers as input to RNNs; when data is scarce, such approaches have difficulty capturing how the hidden representations respond to treatment.

State space models and other Markov models have been used to model the progression of a variety of chronic diseases, including Cystic Fibrosis, scleroderma, breast cancer, COPD and CKD \citep{alaa2019attentive, taghipour2013parameter,wang2014unsupervised, schulam2016integrative,perotte2015risk}. 
There has also been much research in characterizing disease trajectories, subtypes, and correlations between risk factors and progression for patients suffering from Alzheimer's Disease \citep{khatami2019challenges,goyal2018characterizing,zhang2019analysis,marinescu2019dive}. Like us, the above works pose disease progression as density estimation but in contrast, many of the above models do not condition on time-varying interventions. 
\section{Background - State Space Models (SSMs)\label{sec:bg}}

SSMs are a popular model for sequential data and have a rich history in modeling disease progression.

\textbf{Notation:} 
$B \in \R{J}$ denotes baseline data that are static, i.e. individual-specific covariates. For chronic diseases, these data comprise a $J$-dimensional vector, including patients' age, gender, genetics, race, and ethnicity.
Let $\mathbf{U}=\{U_{0}, \ldots, U_{T-1}\};\; U_t\in\R{L}$ be a sequence of $L$-dimensional interventions for an individual. An element of $U_t$ may be binary, to denote prescription of a drug,
or real-valued, to denote dosage.
$\mathbf{X} = \{X_{1}, \ldots, X_{T}\};\;X_t\in\R{M}$ denotes the sequence of real-valued, $M$-dimensional clinical biomarkers. An element of $X_t$ may denote a serum lab value or blood count, which is used by clinicians to measure organ function as a proxy for disease severity.
$X_t$ frequently contains missing data.
We assume access to a dataset $\mathcal{D} = \{(\mathbf{X}^1,\mathbf{U}^1,B^1),\ldots,(\mathbf{X}^N,\mathbf{U}^N,B^N)\}$.
For a visual depiction of the data, we refer the reader to Figure \ref{fig:data_vis}. 
Unless required, we ignore the superscript denoting the index of the datapoint and denote concatenation with [].

\textbf{Model:}
SSMs capture dependencies in sequential data via a time-varying latent state. When this latent state is discrete, SSMs are also known as Hidden Markov Models (HMM). In our setting, we deal with a continuous latent state. The generative process is: 
\begin{align}%
\textstyle
    p(\mathbf{X}|\mathbf{U},B) &= \int_{Z} \prod_{t=1}^{T}p_{\theta}(Z_t|Z_{t-1}, U_{t-1}, B)  p_{\theta}(X_t|Z_t) dZ \nonumber\\
    Z_t | \cdot \sim &\mathcal{N}(\mu_{\theta}(Z_{t-1}, U_{t-1}, B), \Sigma^t_{\theta}(Z_{t-1}, U_{t-1}, B)),\nonumber \\
    X_t | \cdot \sim &\mathcal{N}(\kappa_{\theta}(Z_t),\Sigma^e_{\theta}(Z_t)) \label{eq:ssm_param}
\end{align}
We denote the parameters of a model by $\theta$, which may comprise weight matrices or the parameters of functions that index $\theta$. SSMs make the Markov assumption \emph{on the latent variables}, $Z_t$, and we assume that relevant information about past medications are captured by the state or contained in $U_{t-1}$. 
We set $\Sigma^{t}_{\theta}, \Sigma^{e}_{\theta},\kappa_{\theta}(Z_t)$ to be functions of a concatenation of their inputs, e.g. $\Sigma^{t}_{\theta}(\cdot) = \textbf{softplus}(\mathbf{W}[Z_{t-1},U_{t-1},B] +\mathbf{b})$. $\Sigma^{t}_{\theta}, \Sigma^{e}_{\theta}$ are diagonal matrices where the softplus function is used to ensure positivity.

\textbf{Learning:} We maximize $\sum_{i=1}^N\log p(\mathbf{X}^i|\mathbf{U}^i,B^i)$ with respect to $\theta$. For a nonlinear SSM, this function is intractable, so
we learn via maximizing a variational lower bound on it. To evaluate the bound, we perform probabilistic inference using a structured inference network \citep{krishnan2017structured}. The learning algorithm alternates between predicting variational parameters using a bi-directional recurrent neural network, evaluating a variational upper bound, and making gradient updates jointly with respect to the parameters of the generative model and the inference network.  
When evaluating the likelihood of data under the model, if $X_t$ is missing, it is marginalized out. 
Since the inference network also conditions on sequences of observed data to predict the variational parameters, we use forward fill imputation where data are missing.
\section{Attentive Pharmacodynamic State Space Model \label{sec:methodology}}

\begin{figure*}[t!]
\centering
\resizebox{0.18\textwidth}{!}{
\begin{tikzpicture}
  \node[obs]                               (b) {$B$};
  \node[latent, right=of b,  xshift=-1cm, yshift=-0.8cm]     (x1) {$Z_1$};
  \node[obs, above=of x1, yshift=-0.3cm]     (u1) {$U_1$};
  \node[obs, rectangle, above=of x1, xshift=0.5cm, yshift=-0.9cm,scale=0.3]   (h1) {};
  \node[latent, right=of x1, xshift=-5mm]    (x2) {$Z_2$};
  \node[obs, above=of x2, yshift=-0.3cm]     (u2) {$U_2$};
  \node[obs, rectangle, above=of x2, xshift=0.5cm, yshift=-0.9cm,scale=0.3]   (h2) {};
  \node[latent, right=of x2, xshift=-5mm]    (x3) {$Z_3$};
  \node[obs, below=of x1, yshift=0.7cm]      (y1) {$X_1$};
  \node[obs, below=of x2, yshift=0.7cm]      (y2) {$X_2$};
  \node[obs, below=of x3, yshift=0.7cm]      (y3) {$X_3$};
  \edge {b,u1,x1} {h1} ; %
  \edge {b,u2,x2} {h2} ; %
  \edge {h1} {x2} ; %
  \edge {h2} {x3} ; %
  \edge {x1} {y1} ; %
  \edge {x2} {y2} ; %
  \edge {x3} {y3} ; %
\end{tikzpicture}
}
\includegraphics[width=0.39\textwidth]{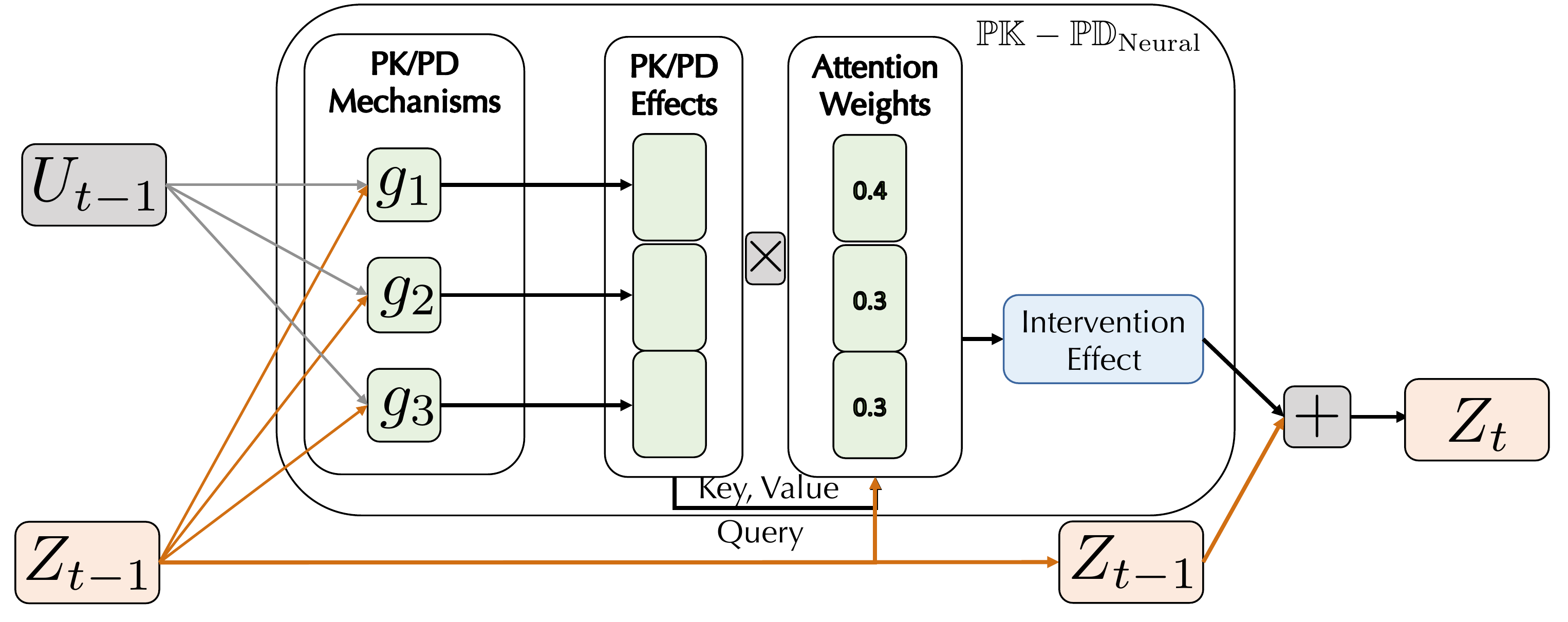}
\includegraphics[width=0.25\textwidth]{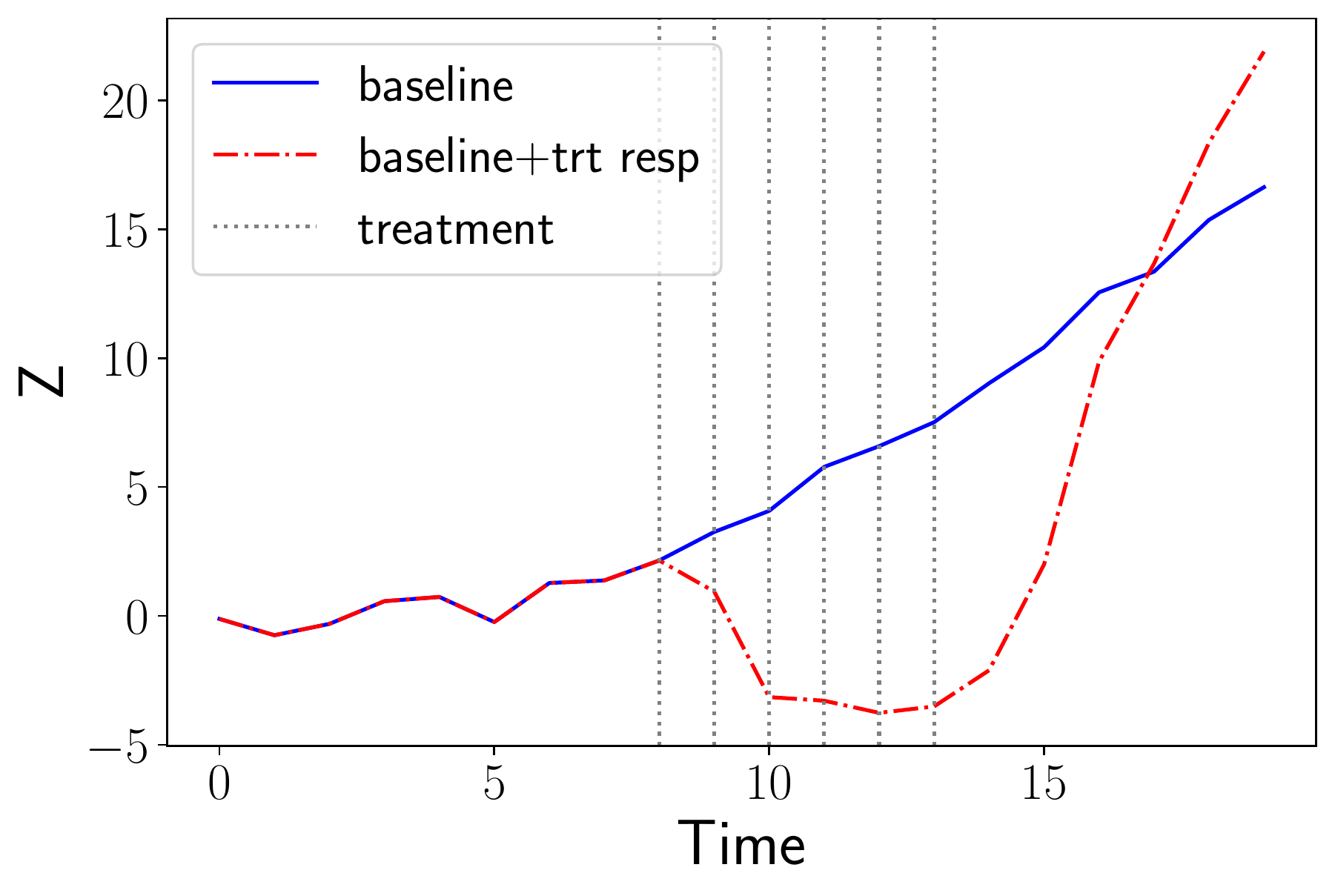}
\caption{\small \textbf{Unsupervised Models of Sequential Data (Left):} We show a State Space Model (SSM) of $\textbf{X}$ (the longitudinal biomarkers) conditioned on $B$ (genetics, demographics) and $\textbf{U}$ (binary indicators of treatment and line of therapy). The rectangle depicts the $\IEF(Z_{t-1},U_{t-1},B)$.
\textbf{Neural Architecture for $\PKPDIEF$ (Middle):} Illustration of the neural architecture we design; we use a soft-attention mechanism over the neural PK/PD effects using the current patient representation as a query to decide
how the masks should be distributed.
 \textbf{Modeling relapse with the neural treatment exponential response (Right):} The curve depicts a single dimension of the representation and vertical lines denote a single treatment. After maintaining the response with treatments, a regression towards baseline (in blue; depicting what would have happened had no treatment been prescribed) occurs when treatment is stopped.
}
\label{fig:data_model}
\end{figure*}
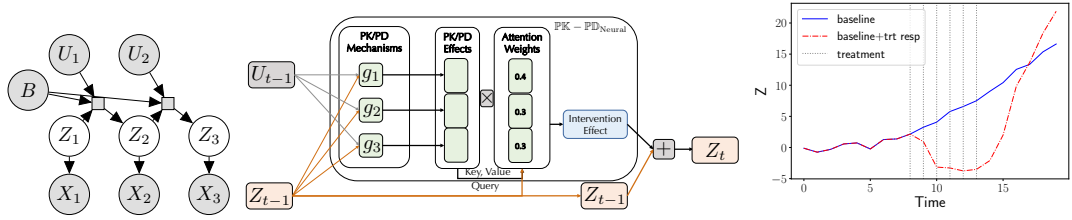

To make the shift from black-box models to those that capture 
useful structure for modeling clinical data, 
we begin with a discussion of PK-PD models and some of the key limitations
that practitioners may face when directly applying them to modern 
clinical datasets.

\subsection{Limitations of Pharmacokinetic-Pharmacodynamic Modeling} 
Pharmacology is a natural store of domain expertise for reasoning about how treatments affect disease. We look specifically at pharmacokinetics (PK), which deals with how drugs move in the body, and pharmacodynamics (PD), which studies the body's response to drugs.
Consider a classical pharmacokinetic-pharmacodynamic (PK-PD) model used to characterize variation in tumor volume due to chemotherapy \cite{norton2014cancer,west2017chemotherapeutic}.
Known as the log-cell kill model, it is based on the hypothesis that a given dose of chemotherapy results in killing a constant fraction of tumor cells rather than a constant number of cells. 
The original model is an ordinary differential equation but an equivalent  expression is:
\begin{equation}
\label{eqn:logcellexample}
    S(t) = S(t-1)\cdot(1 + \rho\log(K/S(t-1))-\beta_c C(t)),
\end{equation}
$S(t)$ is the (scalar) tumor volume, $C(t)$ is the (scalar) concentration of a chemotherapeutic drug over time, $K$ is the maximum tumor volume possible, $\rho$ is the growth rate,
and $\beta_c$ represents the drug effect on tumor size. Besides its bespoke nature, there are some key limitations of this model that hinder its broad applicability for unsupervised learning:

\textit{Single intervention, single biomarker:} The model parameterizes the effect of a \emph{single, scalar} intervention on a single, scalar, time-varying biomarker making it impossible to apply directly to high-dimensional clinical data. Furthermore, the quantity it models, tumor volume, is unobserved for non-solid cancers.

\textit{Misspecified in functional form:} The log-cell-kill hypothesis, by itself, is not an accurate description of the drug mechanism in most non-cancerous chronic diseases.
\textit{Misspecified in time:} Patients go through cycles of recovery and relapse during a disease. Even if the hypothesis holds when the patient is sick, it may not hold when the patient is in recovery.

In what follows, we aim to mitigate these limitations to build a practical, scalable model of disease progression. 

\subsection{Latent Representations of Disease State}
Tackling the first limitation, we use nonlinear SSMs in order to model longitudinal, high-dimensional data.
Even though tumor volume may not be observed in observational clinical datasets, various proxies (e.g. lab values, blood counts) of the unobserved disease burden often are. We conjecture that the time-varying latent representation, $Z_t$, implicitly captures such clinical phenotypes from the observations. 

To ensure that the phenotypes captured by $Z_t$ vary over time in a manner akin to clinical intuition, we focus the efforts of our design on the transition function, $\mu_{\theta}(Z_{t-1}, U_{t-1},B)$, of the state space model. This function controls the way in which the latent state $Z_t$ in an SSM evolves over time (and through it, the data) when exposed to interventions, $U_t$; this makes the transition function a good starting point for incorporating clinical domain knowledge.

\subsection{Neural Attention over Treatment Effect Mechanisms} 
\label{sec:attention_mechanism} 
In order to design a good transition function, we first need to address the second limitation that we may not know the \emph{exact} mechanism by which drugs affect the disease state. However, we often have a set of reasonable hypotheses about the mechanisms that underlie how we expect the dynamics of the latent disease state to behave.

Putting aside the specifics of what mechanisms we should use for the moment, suppose we are given $d$ mechanism functions, $g_1,\ldots,g_d$, each of which is a neural architecture that we believe captures aspects of how a representation should vary as a response to treatment. How a patient's representation should vary will depend on what state the patient is in. e.g. sicker patients may respond less well to treatment than healthier ones. 
To operationalize this insight, we make use of an attention mechanism \citep{bahdanau2014neural} to attend to which choice of function is most appropriate.

\textit{Attending over mechanisms of effect } Attention mechanisms operate by using a "query" to index into a set of "keys" to compute a set of attention
weights, which are a distribution over the "values". We propose a soft-attention mechanism to select between $g_1,\ldots,g_d$.
At each $t$, for the query, we have $q=Z_{t-1}W_q$. For the key and value, we have, 
\begin{align*}
    \Tilde{K} &= [g_1(Z_{t-1},U_{t-1},B);\ldots;g_d(Z_{t-1},U_{t-1},B)]^\top W_k \\
    \Tilde{V} &= [g_1(Z_{t-1},U_{t-1},B);\ldots;g_d(Z_{t-1},U_{t-1},B)]^\top W_v.
\end{align*}
Note that $W_q,W_k,W_v \in \mathbb{R}^{Q \times Q}$ and that $q\in \mathbb{R}^{Q}$, $\Tilde{K}\in \mathbb{R}^{Q \times d}$, and $\Tilde{V} \in \mathbb{R}^{Q \times d}$. Then, we have the following, 
\begin{equation}
\label{eqn:attention_mech}
    \IEF(Z_{t-1},U_{t-1},B) = \bigg(\sum_{i=1}^d \textbf{softmax}\bigg(\frac{q \odot \Tilde{K}}{\sqrt{Q}}\bigg)_i \odot \Tilde{V}_i\bigg)W_o
\end{equation}
We compute the attention weights using the latent representation at a particular time point as a "query" and the output of each of $g_1,\ldots,g_d$ as "keys"; see Figure \ref{fig:data_model} (middle).
This choice of neural architecture for $\mu_{\theta}$ allows us to parameterize 
heterogeneous SSMs, where the function characterizing latent dynamics changes over time. 

\subsection{Lines of Therapy with Local and Global Clocks}\label{sec:local_global_clocks}

Here, we address a third limitation of classical PK-PD models: a proposed drug mechanism's validity may depend on how long the patient has been treated and what stage of therapy they are in. Such stages, or \textit{lines of therapy}, refer to contiguous plans of multiple treatments prescribed to a patient. They are often a unique structure of clinical data from individuals suffering from chronic diseases. 
For example, first line therapies often represent combinations prioritized
due to their efficacy in clinical trials; subsequent lines may be decided by clinician preference.
Lines of therapy index treatment plans that span multiple time-steps and are often laid out by clinicians at first diagnosis.
 We show how to make use of this information within a mechanism function. 

To capture the clinician's intention when prescribing treatment, we incorporate line of therapy as a one-hot vector in  $U_t[:K]\;\forall t$ ($K$ is the maximal line of therapy).
Lines of therapy typically change when a drug combination fails or causes adverse side effects.
By conditioning on line of therapy, a transition function (of the SSM) parameterized by a neural network can, in theory, infer the length of time a patient has been on that line. However, although architectures such as Neural Turing Machines \cite{graves2014neural} can learn to count occurrences, they would need a substantial amount of data to do so.

To enforce the specified drug mechanism functions to capture time since change in line of therapy, we use \emph{clocks} to track the time elapsed since an event. This strategy has precedent in RNNs, where \citet{che2018recurrent} use time since the last observation to help RNNs learn well when data is missing. \citet{koutnik2014clockwork} partition the hidden states in RNNs so they are updated at different time-scales. 
Here, we augment our interventional vector, $U_t$, with two more dimensions. A global clock, $gc$, captures time elapsed since $T=0$, i.e. $U_t[K] = \text{gc}_t = t$.
A local clock, $lc$, captures time elapsed since a line of therapy began; i.e. $U_t[K+1] = \text{lc}_t = t-p_t$ where
$p_t$ denotes the index of time when the line last changed.
By using the local clock, $\IEF(Z_{t-1},U_{t-1},B)$ can modulate $Z_t$ to capture patterns such as: the longer a line of therapy is deployed, the less or (more) effective it may be.

For the patient in Figure \ref{fig:data_vis}, we can see that the first dimension of $\mathbf{U}$ denoting line of therapy would be $[0,0,0,0,1,1,2,2,2]$.
Line $0$ was used four times, line $1$ used twice, line $2$ used thrice. 
Then, $p=[0,0,0,0,4,4,6,6,6,6]$, $gc=[0,1,2,3,4,5,6,7,8,9]$ and $lc = [0,1,2,3,0,1,0,1,2,3]$. 
To the best of our knowledge, we are the first to make use of lines-of-therapy information and clocks concurrently to capture temporal information when modeling clinical data.

\subsection{Neural PK-PD Functions for Chronic Diseases  \label{sec:inductive_biases}}

Having developed solutions to tackle some of the limitations of PK-PD models, we turn to the design of three new mechanism functions, each of which captures different hypotheses a clinician may have about how the underlying disease burden of a patient changes (as manifested in their latent states).

\textbf{Modeling baseline conditional variation: } 
Biomarkers of chronic diseases can increase, decrease, or stay the same. 
Such patterns may be found in the dose-response to chemotherapy used in solid cancerous tumors \citep{klein2009parallel}. 
In reality, clinicians find that these changes are often modulated by patient specific features such as age, genetic mutations, and history of illness. 
Patients who have been in therapy for a long time may find decreased sensitivity to 
treatments. To capture this variation:
\begin{align}
  \textstyle
    &g_1(Z_{t-1},U_{t-1},B) = Z_{t-1}\cdot \text{tanh}(b_{\text{lin}} + W_{\text{lin}}[U_{t-1},B])
\label{eq:lin}
\end{align}
where $b_{\text{lin}}\in\R{Q}, W_{\text{lin}}\in\mathbb{R}^{Q\times(L+J)}$.
Here, the effects on the representation are bounded (via the tanh function) but depend 
on the combination of drugs prescribed and the patient's baseline data, including genetics.

\textbf{Modeling slow, gradual relapse after treatment: } 
One of the defining features of many chronic diseases is the possibility of a relapse during active therapy. 
In cancer, a relapse can happen due to cancerous cells escaping the treatment or a variety of other bio-chemical processes, such as increased resistance to treatment due to mutations. 
The relapse can result in bio-markers
reverting to values that they held prior to the start of treatment; for an example of this, see Figure \ref{fig:data_model} (right). We design the following neural architectures to capture such patterns in a latent representation.

\textit{Neural Log-Cell Kill:} This architecture is inspired by the classical log cell kill model of tumor volume in solid cell tumors \citep{west2017chemotherapeutic} but unlike the original model, scales to high-dimensional representations and 
takes into account \emph{lines of therapy} via the local clock.
This allows the model to effectively reset every time a new
line of therapy begins. The functional form of the model is,
\begin{align}\label{eq:mm_lck}
    &g_2(Z_{t-1},U_{t-1},B) =Z_{t-1}\cdot (1 - \rho\log(Z_{t-1}^2)\\\nonumber
    &-\beta\exp(-\delta \cdot \text{lc}_{t-1})),
\end{align}
where $\beta = \tanh(W_{lc}U_{t-1} + b_{lc})$. $W_{lc}\in \mathbb{R}^{Q\times L}, b_{lc} \in \mathbb{R}^Q, \delta\in \mathbb{R}^Q$ and $\rho\in \mathbb{R}^Q$ are learned. 
While diseases may not have a single observation that characterizes the state of the organ system (akin to tumor volume), we hypothesize that representations, $Z_{t}$, of the observed clinical biomarkers may benefit from mimicking the dynamics exhibited by tumor volume when exposed to chemotherapeutic agents. We emphasize that unlike Equation \ref{eqn:logcellexample}, the function in Equation \ref{eq:mm_lck} operates over a \emph{vector valued} set of representations that can be modulated by the patient's genetic markers.  

\textit{Neural Treatment Exponential:}
\citet{xu2016bayesian} develop a Bayesian nonparameteric model to explain variation in creatinine, a single biomarker, due to treatment.
We design an architecture inspired by their model that scales to high dimensional representations, allows for the representation
to vary as a function of the patient's genetics, and makes use of information in the lines of therapy via the clocks.
\begin{align}\label{eq:trt_exp_main}
    &g_3(Z_{t-1},U_{t-1},B)  \\
    &= \begin{cases}
    &b_0 + \alpha_{1,t-1} / [1+\exp (-\alpha_{2,t-1}(\text{lc}_{t-1}-\frac{\gamma_{l}}{2}))], \nonumber\\ 
    &\text{if } 0 \leq \text{lc}_{t-1} < \gamma_l \\
    &b_l + \alpha_{0,t-1} / [1+\exp (\alpha_{3,t-1}(\text{lc}_{t-1}-\frac{3\gamma_{l}}{2}))], \nonumber\\
    &\text{if } \text{lc}_{t-1} \geq \gamma_l
  \end{cases}
\end{align}

Despite its complexity, the intermediate representations learned within this architecture have simple intuitive meanings. 
$\alpha_{1,t-1} = W_d[Z_{t-1}, U_{t-1}, B] + b_d$, where $W_d\in\mathbb{R}^{Q \times (Q+L+J)}, b_d\in\mathbb{R}^Q$ is used to control whether each dimension in $Z_{t-1}$ increases or decreases as a function of the treatment and baseline data.
$\alpha_{2,t-1}, \alpha_{3,t-1}$, and $\gamma_l$ control the steepness and duration of the intervention effect. We restrict these characteristics to be similar for drugs administered under the same line of therapy. Thus, we parameterize: $[\alpha_2, \alpha_3, \gamma_l]_{t-1} = \sigma(W_e\cdot U_{t-1}[0] + b_e)$. If there are three lines of therapy, $W_e\in\mathbb{R}^{3\times3}, b_e\in\mathbb{R}^3$ and the biases, $b_0\in\mathbb{R}^{Q}$ and $b_l\in\mathbb{R}^{Q}$, are learned. 
Finally, $\alpha_{0,t-1} = (\alpha_{1,t-1} + 2b_0 - b_l) / (1 + \exp(-\alpha_{3,t-1}\gamma_l/2))$ ensures that the effect peaks at $t = \text{lc}_t + \gamma_l$.
Figure \ref{fig:data_model} (right) depicts how a single latent dimension may vary over time for a single line of therapy using this neural architecture. 

\textbf{From $\PKPDIEF$ to the $\SSMpkpd$: } 
When $g_1,g_2,g_3$, as described in Equations \ref{eq:lin}, \ref{eq:mm_lck}, \ref{eq:trt_exp_main}, are used in the transition function $\mu_{\theta}$ (as defined in Equation \ref {eqn:attention_mech}), we refer to the resulting function as $\PKPDIEF$. Moreover, when $\PKPDIEF$ is used as the transition function in an SSM, we refer to the resulting model as $\SSMpkpd$, a heterogeneous state space model designed to model the progression of diseases. 
\section{Evaluation\label{sec:eval}}

\subsection{Datasets}
We study $\SSMpkpd$ on three different datasets -- two here, and on a third semi-synthetic dataset in the appendix.

\textbf{Synthetic Data:}
We begin with a synthetic disease progression dataset where each patient is assigned baseline covariates $B\in\R{6}$. $B$ determines how the biomarkers, $X_{t}\in\R{2}$, behave in the absence of treatment. $U_t\in\R{4}$ comprises the line of therapy ($K=2$), the local clock, and a single binary variable indicating when treatment is prescribed. To mimic the data dynamics described in Figure \ref{fig:data_vis}, the biomarkers follow second-order polynomial trajectories over time with the underlying treatment effect being determined by the Neural Treatment Exponential (see Equation \ref{eq:trt_exp_main}). Biomarker 1 can be thought of as a marker of disease burden, while biomarker 2 can be thought of as a marker of biological function. The full generative process for the data is in the supplementary material. 
To understand generalization of the model as a function of sample complexity, we train on $100$/$1000$ samples and evaluate on five held-out sets of size $50000$.

\textbf{ML-MMRF:} 
The Multiple Myeloma Research Foundation (MMRF) CoMMpass study releases de-identified 
clinical data for $1143$ patients suffering from multiple myeloma, an incurable plasma cell cancer. All patients are aligned to the start of treatment, which is made according to current standard of care (not random assignment). 
With an oncologist, we curate demographic and genomic markers, $B\in\R{16}$, clinical biomarkers, $X_t\in\R{16}$, and interventions, $U_t\in\R{9}$, with one local clock, a three dimensional one-hot encoding for line of therapy, and binary markers of $5$ drugs. 
Our results are obtained using a 75/25 train/test split. To select hyperparameters, we perform 5-fold cross validation on the training set. 
Finally, there is missingness in the biomarkers, with $66\%$ of the observations missing. We refer the reader to the appendix for more details on the dataset.

\subsection{Setup} 
We learn via: ($\arg\min_{\theta} -\log p(\mathbf{X}|\mathbf{U},B;\theta)$) using ADAM \citep{kingma2014adam} with a learning rate of $0.001$ for $15000$ epochs. L$1$ or L$2$ regularization is applied in one of two ways: either we regularize all model parameters (including parameters of inference network), or we regularize all weight matrices except those associated with the attention mechanism. We search over regularization strengths of $0.01,0.1,1,10$ and latent dimensions of $16, 48, 64$ and $128$. We do model selection using the negative evidence lower bound (NELBO); Appendix B contains details on the derivation of this bound. We use a single copy of $g_1$, $g_2$, and $g_3$ in the transition function. Multiple copies of each function as well as other "mechanistic" functions can be used, highlighting the flexibility of our approach. However, this must be balanced with potentially overfitting on small datasets.

\subsection{Baselines} 
$\SSMlinear$ parametrizes $\mu_{\theta}(Z_{t-1}, U_{t-1},B)$ with a linear function. This model is a strong, linear baseline whose variants have been used for modeling data of patients suffering from Chronic Kidney Disease \citep{perotte2015risk}.

$\SSMnl$: \citet{krishnan2017structured} use a nonlinear SSM to capture variation in the clinical biomarkers of diabetic patients. We compare to their model, parameterizing the transition function with a 2-layer MLP.

$\SSMmoe$: We use an SSM whose transition function is 
parameterized via a Mixture-of-Experts (MoE) architecture \citep{jacobs1991adaptive,jordan1994hierarchical}; i.e. $g_1,g_2,g_3$ are each replaced with a multi-layer perceptron.
This baseline does not incorporate any domain knowledge and tests the relative benefits of prescribing the functional forms via mechanisms versus learning them from data. 

$\textbf{SSM}_{\text{Attn.Hist.}}$: We implement a variant of the SSM in \citet{alaa2019attentive}, a state-of-the-art model for disease progression trained via conditional density estimation. The authors use a \textit{discrete} state space for disease progression modeling making a direct comparison difficult.
However, $\textbf{SSM}_{\text{Attn.Hist.}}$ preserves the structural modeling assumptions they make. Namely, the transition function of the model attends to a concatenation of previous states and interventions at each point in time. We defer specifics to Appendix B.

In addition, we run two simpler baselines, a First Order Markov Model (FOMM) and Gated Recurrent Unit (GRU) \citep{cho2014properties}, on the synthetic data and ML-MMRF but defer those results to Appendix E.

\subsection{Evaluation Metrics\label{sec:metrics}} 
\textbf{NELBO } On both the synthetic data and ML-MMRF data, we quantify generalization via the negative evidence lower bound (NELBO), which is a variational upper bound on the negative log-likelihood of the data. A lower NELBO indicates better generalization. 

\textbf{Pairwise Comparisons } For a fine-grain evaluation of our models on ML-MMRF, we compare held-out NELBO under $\SSMpkpd$ versus the corresponding baseline for each patient. For each held-out point, $\Delta_i = 1$ when the NELBO of that datapoint is lower under $\SSMpkpd$ and $\Delta_i = 0$ when it is not. 
In Table \ref{tab:tabglob} (bottom), we report $\frac{1}{N}\sum_{i=1}^N \Delta_i$, the proportion of data for which $\SSMpkpd$ yields better results. 

\textbf{Counts } To get a sense for the number of patients on whom $\SSMpkpd$ does much better, we count the number of held-out patients for whom the held-out negative log likelihood (computed via importance sampling) is more than $10$ nats lower under $\SSMpkpd$ than the corresponding baseline (and vice versa for the baselines).

\begin{table*}[t!]
\centering
\scalebox{0.8}{
\begin{tabular}[t]{||c| c c c c c c||}
\toprule
    \begin{tabular}[c]{@{}c@{}}\textbf{Training Set Size}\end{tabular} &
 \textbf{\begin{tabular}[c]{@{}c@{}}SSM\\ Linear\end{tabular}} &
 \textbf{\begin{tabular}[c]{@{}c@{}}SSM\\ NL\end{tabular}} &
 \textbf{\begin{tabular}[c]{@{}c@{}}SSM\\ MOE\end{tabular}} & 
 \textbf{\begin{tabular}[c]{@{}c@{}}SSM\\ Attn. Hist.\end{tabular}} &
 \textbf{\begin{tabular}[c]{@{}c@{}}SSM\\ PK-PD\end{tabular}} &  \textbf{\begin{tabular}[c]{@{}c@{}}SSM PK-PD \\(w/o TExp)\end{tabular}}\\
 \midrule
100 & 58.57 +/- .06 & 69.04 +/- .11 & 60.98 +/- .04 & 76.94 +/- .02 & \textbf{55.34 +/- .03} & \textbf{58.39 +/- .05}\\
 \midrule 
1000 & 53.84 +/- .02 & 44.75 +/- .02 & 51.57 +/- .03 & 73.80 +/- .03 & \textbf{39.84 +/- .02} & \textbf{38.93 +/- .01}\\ 

\bottomrule
\end{tabular}
}
\scalebox{0.8}{
 \begin{tabular}[t]{||c | c c c c||}
    \toprule
      \textbf{\begin{tabular}[c]{@{}c@{}}Evaluation  Metric\end{tabular}} & \textbf{\begin{tabular}[c]{@{}c@{}}SSM\\ PK-PD \end{tabular}} vs.\textbf{ \begin{tabular}[c]{@{}c@{}}SSM\\ Linear\end{tabular}} &  \textbf{\begin{tabular}[c]{@{}c@{}}SSM\\ PK-PD \end{tabular}} vs.\textbf{\begin{tabular}[c]{@{}c@{}}SSM\\ NL\end{tabular}} &
     \textbf{\begin{tabular}[c]{@{}c@{}}SSM\\ PK-PD \end{tabular}} vs. \textbf{\begin{tabular}[c]{@{}c@{}}SSM\\ MOE\end{tabular}} & 
     \textbf{\begin{tabular}[c]{@{}c@{}}SSM\\ PK-PD \end{tabular}} vs. \textbf{\begin{tabular}[c]{@{}c@{}}SSM\\ Attn. Hist. \end{tabular}}\\
    \midrule                                                    
     Pairwise Comparison ($\uparrow$)  & 0.796 (0.400)     &  0.760 (0.426)    & 0.714 (0.450) & 0.934 (0.247)\\ 
     Counts ($\uparrow$)  & \text{PK-PD}: 158, \text{LIN}: 6      & 130,  12       &  94, 8 &  272, 0\\
     NELBO ($\downarrow$)   & \text{PK-PD}: 61.54, \text{LIN}: 74.22       & 61.54, 79.10       & 61.54, 73.44  & 61.54, 105.04 \\
     \midrule
     $\#$ of Model Parameters   & \text{PK-PD}: 23K, \text{LIN}: 7K        & 23K, 51K       & 23K, 77K  & 23K, 17K\\
    \bottomrule
    \end{tabular}
}
\caption{\small \emph{Top:} \textbf{Synthetic data:} Lower is better. We report the test NELBO with std. dev. for each SSM model to study generalization in the synthetic setting. \emph{Bottom:} \textbf{ML-MMRF:} \textit{Pairwise Comparison}: each number is the fraction (with std. dev.) of test patients for which $\SSMpkpd$ has a lower NELBO than an SSM that uses a different transition function. \textit{Counts}: We report the number of test patients (out of 282) for which an SSM model (PK-PD or otherwise) has a greater than 10 nats difference in negative log likelihood compared to the alternative model. \textit{NELBO}: We report the test NELBO of each model. Note that we label the metrics associated with $\SSMpkpd$ and $\SSMlinear$ in the first column, but drop these labels in subsequent columns.
}
\label{tab:tabglob}
\end{table*}
\subsection{Results}
We investigate three broad categories of questions. 
\subsubsection{Generalization under different conditions}
\textit{$\SSMpkpd$ generalizes better in setting with few ($\sim100$) samples.}
Table \ref{tab:tabglob} (top) depicts NELBOs on held-out synthetic data 
across different models, where a lower number implies better generalization. 
We see a statistically significant gain in generalization for $\SSMpkpd$ compared to all other baselines. $\SSMnl$, $\SSMmoe$ overfit quickly on $100$ samples but recover their performance when learning with $1000$ samples. 

\textit{$\SSMpkpd$ generalizes well when it is misspecified.}
Because we often lack prior knowledge about the true underlying dynamics in the data, we study how $\SSMpkpd$ performs when it is misspecified. We replace the Neural Treatment Exponential function, $g_3$, from $\PKPDIEF$ with another instance of $g_1$. The resulting model is now misspecified since $g_3$ is used to generate the data but no longer lies within the model family. 
We denote this model as (SSM PK-PD w/o TExp). 
In Table \ref{tab:tabglob} (top), when comparing the sixth column to the others, we find that we outperform all baselines and get comparable generalization to $\SSMpkpd$ with the Neural Treatment Exponential function. This result emphasizes our architecture's flexibility and its ability to learn the underlying (unknown) intervention effect through a combination of other, related mechanism functions.

\textit{$\SSMpkpd$ generalizes well on real-world patient data.} A substantially harder test of model misspecification is on the ML-MMRF data where we have unknown dynamics that drive the high-dimensional (often missing) biomarkers in addition 
to combinations of drugs prescribed over time. To rigorously validate whether we improve generalization on ML-MMRF data with $\SSMpkpd$, we study model performance with respect to the three metrics introduced in Section \ref{sec:metrics}. We report our results in Table \ref{tab:tabglob} (bottom). First, we consistently observe that a high fraction of patient data in the test set are explained better by $\SSMpkpd$ than the corresponding baseline (pairwise comparisons). We also note that out of $282$ patients in the test set, across all the baselines, we find that the $\SSMpkpd$ generalizes better for many more patients (counts). Finally, $\SSMpkpd$ has lower NELBO averaged across the entire test set compared to all baselines. 

\begin{figure}
\includegraphics[width=0.43\textwidth]{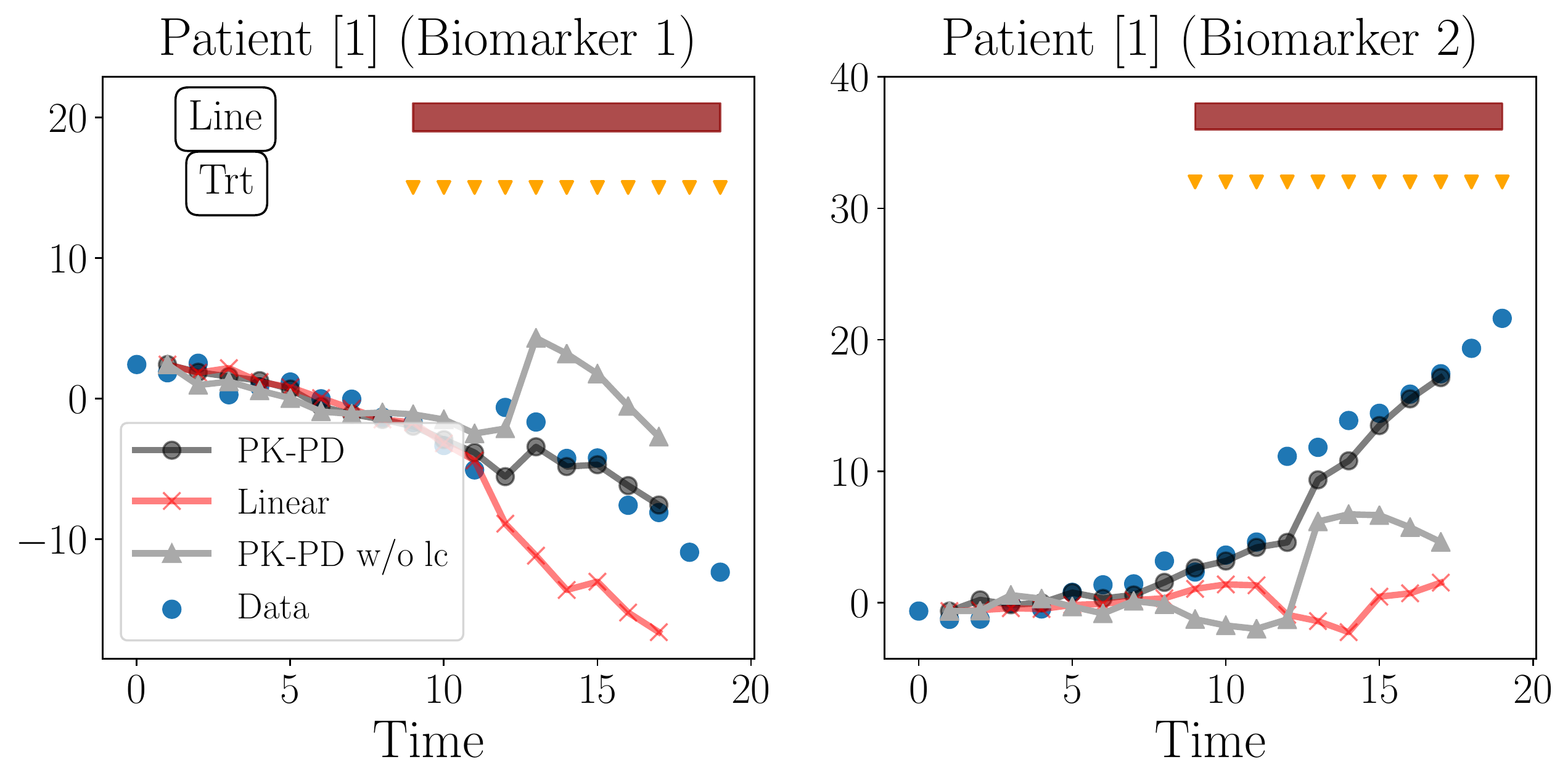}
\centering
\caption{\small \emph{Synthetic}: Forward samples (conditioned only on $B$) from $\SSMpkpd$ (\textcolor{black}{o}), $\SSMlinear$ (\textcolor{red}{x}), $\SSMpkpd$ without local clocks (\textcolor{gray}{$\triangle$}), for a single patient. Y-axis shows biomarker values.}
\label{fig:syn_samples}
\end{figure}

\subsubsection{Model complexity \& generalization}
\textit{The improvements of $\SSMpkpd$ are consistent taking model sizes into account.}
We show in Table \ref{tab:tabglob} (bottom) the number of parameters used in each model. We find that more parameters do not imply better performance. Models with the most parameters (e.g. $\SSMnl$) overfit while those with the lowest number of parameters underfit (e.g. $\SSMlinear$) suggesting that the gains in generalization that we observe are coming from our parameterization.
We experimented with increasing the size of the $\SSMlinear$ model (via the latent variable dimension) to match the size of the best PK-PD model. We found that doing so \emph{did not outperform} the held-out likelihood of $\SSMpkpd$. 

\textit{When data are scarce, a Mixture of Experts architecture is difficult to learn:}
How effective are the functional forms of the neural architectures we develop? %
To answer this question, we compare
the held-out log-likelihood of $\SSMpkpd$ vs $\SSMmoe$ in the third column of Table \ref{tab:tabglob} (bottom).
In the ML-MMRF data, we find that the $\SSMpkpd$
outperforms the $\SSMmoe$. We suspect this is due to the fact that learning diverse "experts" is
hard when data is scarce and supports the hypothesis that the judicious choice of neural architectures
plays a vital role in capturing biomarker dynamics. 

\textit{Can $\PKPDIEF$ be used in other model families?} In the supplement, we
implement $\PKPDIEF$ in a first-order Markov model and find similar
improvements in generalization on the ML-MMRF dataset. This result suggests that the principle we propose of leveraging domain knowledge from pharmacology to design mechanism functions can allow other kinds of deep generative models (beyond SSMs) to also generalize better when data are scarce. 

\begin{figure}
\centering
\includegraphics[width=0.30\textwidth]{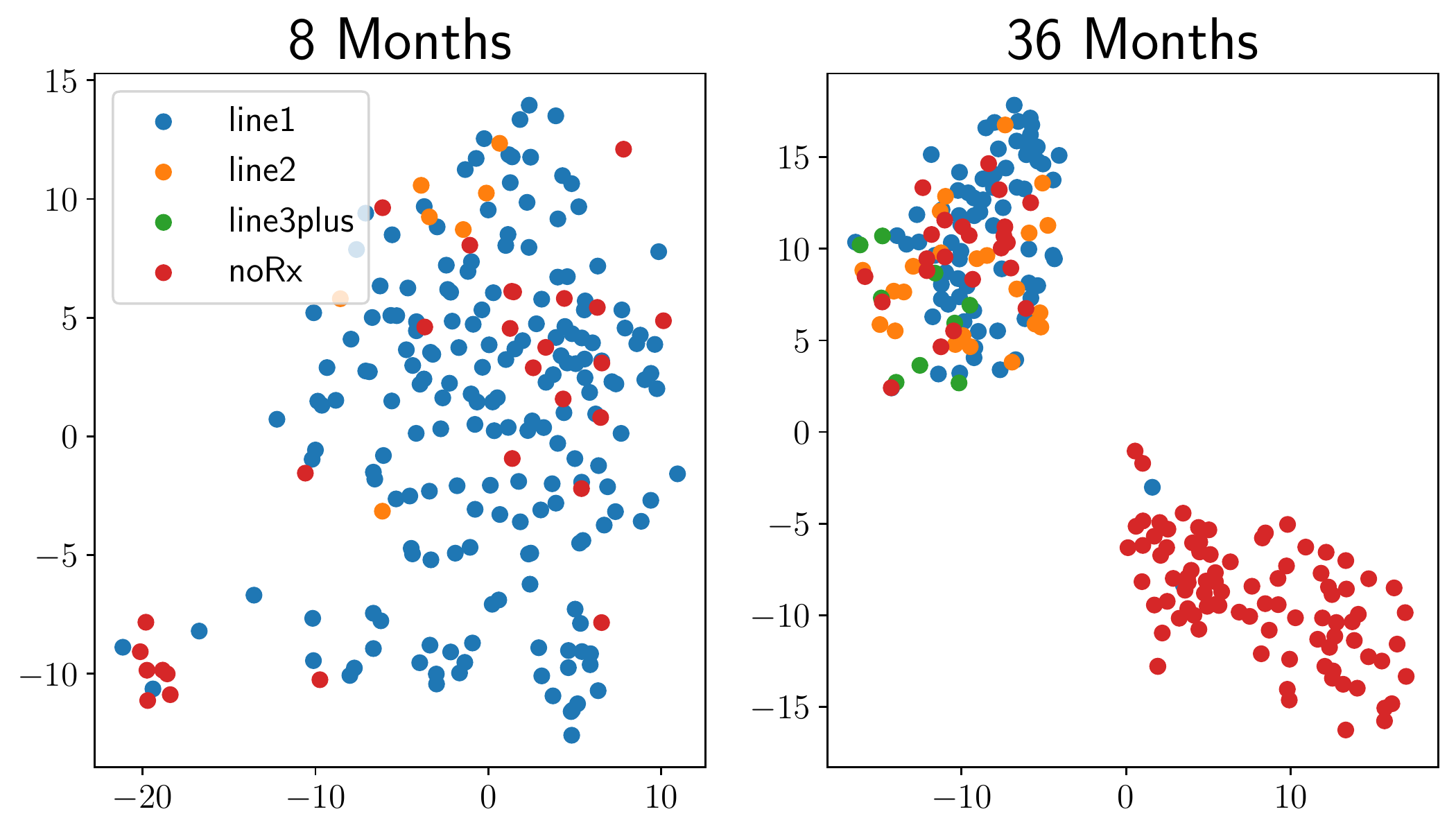}
\includegraphics[width=0.16\textwidth]{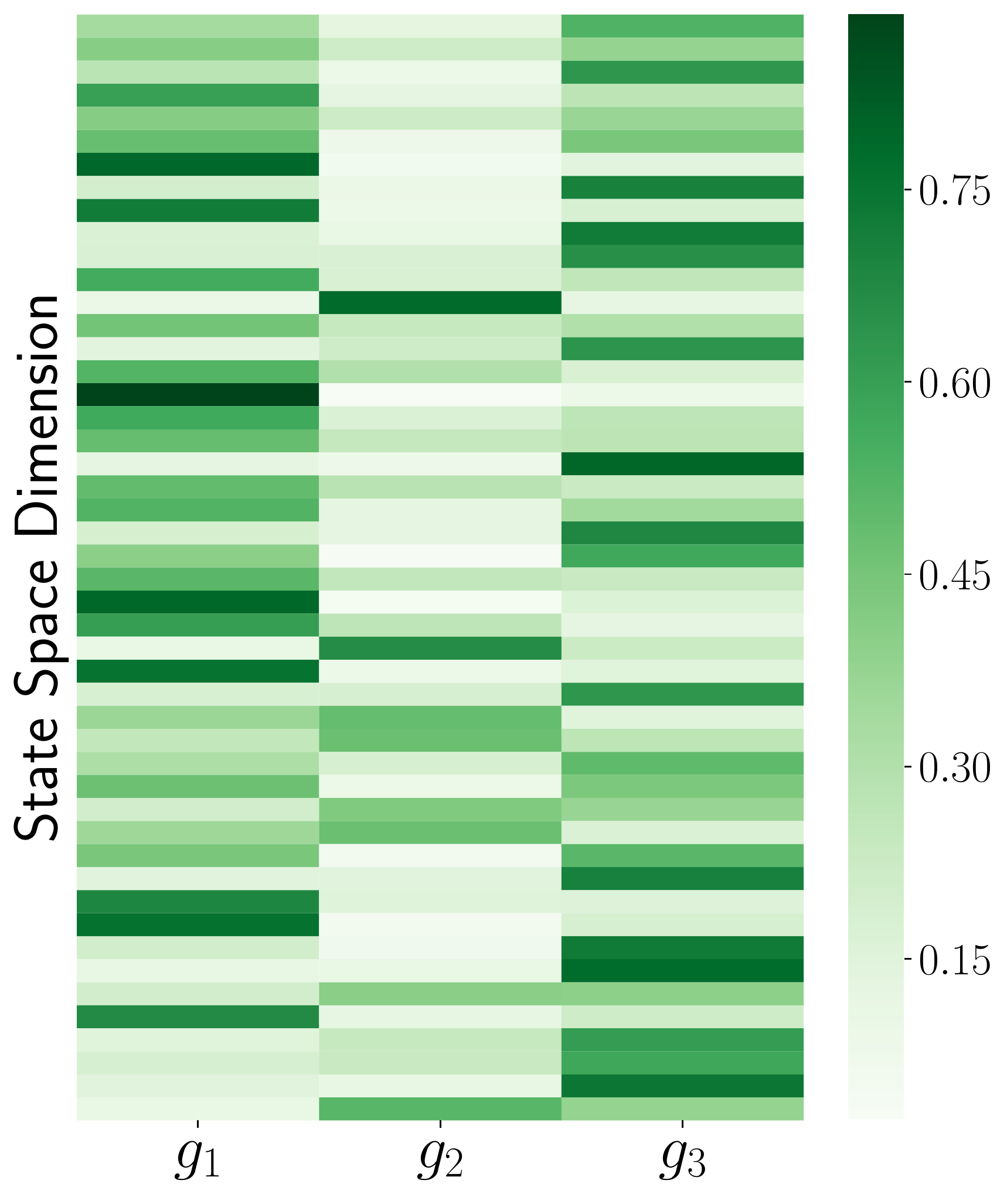}
\caption{\small \emph{ML-MMRF}: (Left two) We visualize the TSNE representations of each test patient's latent state, $Z_t$, at the start of treatment and three years in.
(Right) For $\textbf{SSM}_\text{PK-PD}$, we visualize the attention weights, averaged over all time steps and all patients, on each 
of the neural effect functions across state space dimensions.}
\label{fig:lat_space_attention}
\end{figure}

\subsubsection{Visualizing Patient Dynamics} 
In Figure \ref{fig:lat_space_attention} (right), to further validate our initial hypothesis that the model is using the various neural PK-PD effect functions, we visualize the attention weights from $\SSMpkpd$ trained on ML-MMRF averaged across time and all patients. The highest weighted component is the treatment exponential model $g_3$, followed by the bounded linear model $g_1$ for many of the latent state dimensions. 
We also see that several of the latent state dimensions make exclusive use of the neural log-cell kill model $g_2$. 

\textit{How do the clocks help model patient dynamics? } 
Figure \ref{fig:syn_samples} shows samples from three SSMs trained on synthetic data. $\SSMpkpd$ captures treatment response accurately while $\SSMlinear$ does not register that the effect of treatment can persist over time. To study the impact of clocks on the learned model, we perform an ablation study on SSMs where the local clock in $U_t$, used by $\PKPDIEF$, is set to a constant. Without clocks (PK-PD w/o lc), the model does not capture the onset or persistence of treatment response.

\begin{figure}
\centering
\includegraphics[width=0.41 \textwidth]{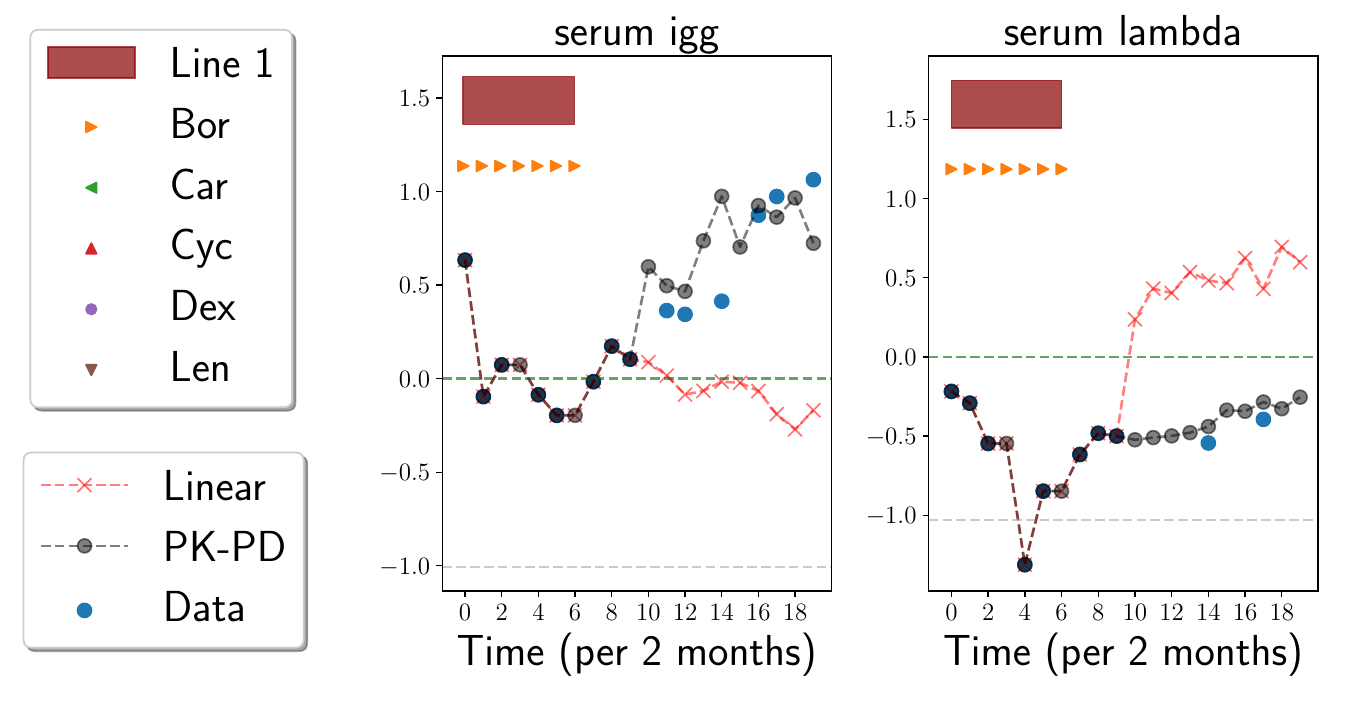}
\caption{\small \emph{ML-MMRF}: Each column is a different biomarker containing forward samples (conditioned on approximately the first 2 years of a patient's data) from $\textbf{SSM}_\text{PK-PD}$ (\textcolor{black}{o}) and $\textbf{SSM}_\text{linear}$ (\textcolor{red}{x}) of a single test patient. Blue circles denote ground truth, and the markers above the trajectories represent treatments prescribed across time. Y-axis shows biomarker levels, with the dotted green[gray] line representing the maximum[minimum] healthy value. Car, Cyc, Dex, and Len shown in legend to maintain consistency with plots in Appendix, but are not given in the treatment regimen.}
\label{fig:mm_samples}
\end{figure}
\textit{$\SSMpkpd$ learns latent representations that reflect the patient's disease state:}
In ML-MMRF, we restrict the patient population to those with at least $T=36$ months of data. At two different
points during their treatment of the disease, we visualize the result of 
TSNE \citep{maaten2008visualizing} 
applied to their latent representations in Figure \ref{fig:lat_space_attention} (left). 

Early in their treatment, the latent representations of these patients appear to have no apparent structure. As time progresses, we find that the dimensions split into two groups. One group, for the most part, is still being treated, while the other is not being treated. A deeper dive into the untreated patients reveals that this cohort has a less severe subtype of myeloma (via a common risk assessment method known as ISS staging). This result suggests that the latent state of $\SSMpkpd$ has successfully 
captured the coarse disease severity of patients at particular time points.

\textit{Visualizing patient samples from $\SSMpkpd$:}
Figure \ref{fig:mm_samples} shows the average of three samples from $\SSMlinear$ and $\SSMpkpd$ trained on ML-MMRF. We track two biomarkers used by clinicians to map myeloma progression. $\SSMpkpd$ better captures the evolution of these biomarkers conditioned on treatment. For serum IgG, $\SSMpkpd$ correctly predicts the relapse of disease after stopping first line therapy, while $\SSMlinear$ does not. On the other hand, for serum lambda, $\SSMpkpd$ correctly predicts it will remain steady.
\section{Discussion\label{sec:discussion}}

$\PKPDIEF$ leverages domain knowledge from pharmacology in the form of treatment effect mechanisms to quantitatively and qualitatively improve performance of a representation-learning based disease progression model. 
\citet{bica2020real} note the potential for blending ideas from 
pharmacology with machine learning: our work is among the first to do so.

We highlight several avenues for future work. 
For example, in machine learning for healthcare, the model we develop can find use in practical problems such as forecasting high-dimensional clinical biomarkers and learning useful representations of patients that are predictive of outcomes.
Doing so can aid in the development of tools for risk stratification or in software
to aid in clinical trial design. 
Applying the model to data from other chronic diseases such as diabetes, congestive heart failure, small cell lung cancer and rheumatoid arthritis, brings opportunities to augment $\PKPDIEF$ with new neural PK-PD functions tailored to capture the unique biomaker dynamics exhibited by patients suffering from these diseases. 

Finally, we believe $\PKPDIEF$ can find use in the design of parameteric environment simulators for different domains. In pharmacology, such simulation based pipelines can help determine effective drug doses \citep{hutchinson2019models}.
Our idea of attending over multiple dynamics functions can find use in the design of simulators in domains such as economics, where multiple hypothesized mechanisms are used to explain observed market phenomena \citep{ghosh2019analysis}.

\section*{Acknowledgements}
The authors thank Rebecca Boiarsky, Christina Ji, Monica Agrawal, Divya Gopinath for valuable feedback on the manuscript and many helpful discussions; Dr. Andrew Yee (Massachusetts General Hospital)
for help in the construction of the ML-MMRF dataset; and both Rebecca Boiarsky and Isaac Lage for their initial analyses and processing of data from the CoMMpass study. These data were generated as part of the Multiple Myeloma Research Foundation Personalized Medicine Initiatives (\url{https://research.themmrf.org} and \url{www.themmrf.org}). This research was generously supported by an ASPIRE award from The Mark Foundation for Cancer Research.

\bibliography{refs.bib}
\bibliographystyle{icml2021}

\clearpage
\appendix 
\section*{Supplementary Material}

The supplementary material contains the following sections. For each section, we highlight the key findings about the experiments we conduct. 
\begin{enumerate}[label=\Alph*.]
    \item \textbf{Learning Algorithms}: This section expands upon the learning algorithm for $\SSMpkpd$ in the main paper. We also describe two additional sequential models -- a First Order Markov Model (FOMM) and a Gated Recurrent Neural Network (GRU). %
    \item \textbf{Synthetic Dataset}: This section provides an in-depth description of the generative process that underlies the synthetic dataset used in the experimental section.
    \item \textbf{The Multiple Myeloma Research Foundation CoMMpass Study}: This section provides details on data extraction, pre-processing and construction of the ML-MMRF dataset.
    \item \textbf{FOMM and GRU Experiments}: We study how incorporating a variant of $\PKPDIEF$ into a FOMM and GRU improves model generalization. The key take-away from this section, with supporting evidence in Table \ref{tab:tabglobapp1p}, is that $\SSMpkpd$ improves generalization not just in state space models, but also in other popular choices of neural network based models of sequences such as FOMMs and GRUs. 
    \item \textbf{Semi-synthetic Experiments}: We introduce a semi-synthetic dataset that we use to further evaluate $\SSMpkpd$. The key take-away from this section, with supporting evidence in Table \ref{tab:tabglob_semi}, is that $\SSMpkpd$ improves generalization on a new dataset whose sequential patterns mimic real-world multiple myeloma data. These improvements are confirmed in a model misspecification scenario.
    \item \textbf{Additional Experiments}: This section details additional experiments to interpret the model we develop and understand the relative utility of its various parts. 
        \begin{enumerate}[label=F\arabic*.]
            \item \textbf{Patient Forecasting} -  We explore different ways in which $\SSMpkpd$ may be used to forecast patient trajectories given some initial data. When conditioning on different lengths of patient history and then sampling forward in time, we see a qualitative improvement in samples from $\SSMpkpd$ compared to one of the best performing baselines.
            \item \textbf{Visualizing Disease Progression} - We extend our analysis of the $\SSMpkpd$'s latent states to studying how they evolve over the entire disease course. We find that clustering patients based on the latent state reveals subgroups that, due to differences in disease severity, have been assigned different treatment regimens. This result suggests that the latent representation has encoded the patient's underlying disease state.
            \item \textbf{Per-feature Breakdown} - We perform a per-feature analysis of how well $\SSMpkpd$ and $\SSMlinear$ model different clinical biomarkers, finding that $\SSMpkpd$ does particularly well for important markers of progression, such as serum IgA.
            \item \textbf{Ablation Analysis} - We study which treatment mechanism function yields the most benefit for modeling the ML-MMRF dataset. Our analysis finds that the Neural Treatment Exponential function provides the most differential gains in NELBO and that the time-varying treatments are crucial for accurately modeling the dynamics of serums IgA, IgG, and Lambda.
        \end{enumerate}

\end{enumerate}

\newpage 
\section{Learning Algorithms \label{sec:app_learning}}
We implement all the models that we experiment with in PyTorch \cite{NEURIPS2019_9015}. 

\paragraph{State Space Models} Recall that the generative process is: 
\begin{align}
\textstyle
    &p(\mathbf{X}|\mathbf{U},B) = \int_{Z} \prod_{t=1}^{T}p(Z_t|Z_{t-1}, U_{t-1}, B;\theta)  p(X_t|Z_t;\theta) dZ\nonumber\\
    &Z_t|\cdot \sim \mathcal{N}(\mu_{\theta}(Z_{t-1}, U_{t-1}, B), \Sigma^t_{\theta}(Z_{t-1}, U_{t-1}, B)),\;\;\nonumber\\
    &X_t|\cdot \sim \mathcal{N}(\kappa_{\theta}(Z_t),\Sigma^e_{\theta}(Z_t)) \nonumber
\end{align}
where the transition function, $\mu_{\theta}$, differs as described in the main paper for $\SSMlinear,\SSMnl,\SSMpkpd \& \SSMmoe$. 

\textit{Maximum Likelihood Estimation of $\theta$:}
Since the log likelihood $ p(\mathbf{X}|\mathbf{U},B)$ is difficult to evaluate and maximize directly 
due to the high-dimensional integral, we resort to a variational learning algorithm that instead maximizes a lower bound on the log-likelihood to learn the model parameters, $\theta$. 
We make use of a structured inference network \cite{krishnan2017structured} that amortizes the
variational approximation, $q_{\phi}(\mathbf{Z}|\mathbf{X})$, to the posterior distribution, $p_{\theta}(\textbf{Z}|\textbf{X})$, of each datapoint.

\begin{align}\label{eq:vlb_ssm}
    &\log p(\mathbf{X}|\mathbf{U},B;\theta)\geq\mathcal{L}(\mathbf{X}; (\theta, \phi)) \\
    &= \sum_{t=1}^{T}\mathbb{E}_{q_{\phi}(Z_t|\mathbf{X},\mathbf{U},B)} [\log p_{\theta}(X_t|Z_t)]\nonumber\\
    &- \text{KL}(q_{\phi}(Z_1|\mathbf{X},\mathbf{U},B)||p_{\theta}(Z_1|B)) \nonumber \\
    &- \sum_{t=2}^T \mathbb{E}_{q_{\phi}(Z_{t-1}|\mathbf{X},\mathbf{U},B)} [\nonumber\\
    &\text{KL}(q_{\phi}(Z_t|Z_{t-1},\mathbf{X},\mathbf{U})||p_{\theta}(Z_t|Z_{t-1}, U_{t-1}, B))] \nonumber
\end{align}

The lower bound on the log-likelihood of data, $\mathcal{L}(\mathbf{X}; (\theta, \phi))$, is a differentiable function of the parameters $\theta,\phi$ \cite{krishnan2017structured}, so we jointly learn them via gradient ascent. As mentioned in the main paper, if $X_t$ is missing, then it is marginalized out when evaluating the likelihood of the data under the model. Since the inference network also conditions on sequences of observed data to predict the variational parameters, we use forward fill imputation where data are missing.

\textit{Hyperparameters:} We present the results of the hyperparameter search on the datasets that we study. Please see the evaluation section of the main paper for the specific ranges that we searched over.
\begin{itemize}
    \item $\SSMlinear$
    \begin{enumerate}
        \item \textit{Synthetic:} State space dimension $48$, L$2$ regularization on all parameters with strength $0.01$
        \item \textit{ML-MMRF:} State space dimension $16$, L$2$ regularization on all parameters with strength $0.01$
    \end{enumerate}
    \item $\SSMnl$
    \begin{enumerate}
        \item \textit{Synthetic:} State space dimension $48$, hidden layer dimension $300$, L$2$ regularization on all parameters with strength $0.1$
        \item \textit{ML-MMRF:} State space dimension $48$, hidden layer dimension $300$, L$2$ regularization on all parameters with strength $0.1$
    \end{enumerate}
    \item $\SSMpkpd$
    \begin{enumerate}
        \item \textit{Synthetic:} State space dimension $48$, L$1$ regularization on subset of parameters with strength $0.01$
        \item \textit{ML-MMRF:} State space dimension $48$, L$1$ regularization on all parameters with strength $0.01$
    \end{enumerate}
    \item $\SSMmoe$
    \begin{enumerate}
        \item \textit{Synthetic:} State space dimension $16$, hidden layer dimension $300$, L$1$ regularization on all parameters with strength $0.01$ 
        \item \textit{ML-MMRF:} State space dimension $48$, hidden layer dimension $300$, L$1$ regularization on all parameters with strength $0.01$ 
    \end{enumerate}
    \item \textbf{SSM Attn. Hist.}
    \begin{enumerate}
        \item \textit{Synthetic:} State space dimension $16$, hidden layer dimension $100$, L$1$ regularization on all parameters with strength $0.01$ 
        \item \textit{ML-MMRF:} State space dimension $48$, hidden layer dimension $300$, L$1$ regularization on all parameters with strength $0.01$ 
    \end{enumerate}
\end{itemize}
\paragraph{SSM Attn. Hist. Baseline:} We provide details on the SSM architecture proposed by \citep{alaa2019attentive} for disease progression modeling. The generative process of their architecture differs from a normal state space model in that the transition function, $\mu_{\theta}$, assumes that the patient's latent state at time $t$ depends on their entire history of latent states and interventions. Thus, we have, 
\begin{align}
\textstyle
    &p(\mathbf{X}|\mathbf{U},B) = \\
    &\int_{Z} \prod_{t=1}^{T}p(Z_t|Z_{1:t-1}, U_{1:t-1}, B;\theta)p(X_t|Z_t;\theta) dZ\nonumber\\
    &Z_t|\cdot \sim \mathcal{N}(\mu_{\theta}(Z_{1:t-1}, U_{1:t-1}, B), \Sigma^t_{\theta}(Z_{1:t-1}, U_{1:t-1}, B)),\;\;\nonumber\\
    &X_t|\cdot \sim \mathcal{N}(\kappa_{\theta}(Z_t),\Sigma^e_{\theta}(Z_t)) \nonumber
\end{align}
Note that we adapt the authors' model to work with a continuous latent state, whereas they utilize a discrete latent state. The crux of their method is to parameterize the transition distribution as an attention-weighted sum of the previous latent states to compute the current latent state. These attention weights are a function of a patient's entire clinical lab and treatment history. Therefore, the transition function that we use to capture their modeling assumptions is as follows:
\begin{equation}
    \mu_{\theta}(Z_{1:t-1},\boldsymbol{\alpha}_{1:t-1}) = W_h(\sum_{i=1}^{t-1}\boldsymbol{\alpha}_i \odot Z_i) + b_h,
\end{equation}
where $\boldsymbol{\alpha}_{1:t-1} = A_t([X_{1:t-1},U_{1:t-1}])$ via an attention mechanism, $A_t$. We use a bi-directional recurrent neural network for the inference network, as opposed to the authors' proposed attentive inference network. We argue that the bi-RNN is just as expressive, since the variational parameters are a function of all past and future observations. Moreover, our goal is to study the effect of altering the generative model in this work.

We also experiment with First Order Markov Models (\textbf{FOMM}) and Gated Recurrent Units (\textbf{GRU}) \citep{chung2014empirical}, which we detail below. 
\paragraph{First Order Markov Models} FOMMs assume observations are conditionally independent of the past given the previous observation, intervention and baseline covariates. The generative process is: 
\begin{align}%
\textstyle
    &p(\mathbf{X}|\mathbf{U},B) = \prod_{t=1}^{T}p(X_t|X_{t-1}, U_{t-1}, B);\;\nonumber\\
    &X_t|\cdot \sim \mathcal{N}(\mu_{\theta}(X_{t-1}, U_{t-1},B), \Sigma_{\theta}(X_{t-1},U_{t-1}, B)),\nonumber
\end{align}
where the transition function, $\mu_{\theta}$, differs akin to the transition function of \textbf{SSM} models, as described in the main paper. Here, we will experiment with $\FOMMlinear, \FOMMnl, \FOMMmoe, \& \FOMMpkpd$

\textit{$\PKPDIEF$ for $\FOMMpkpd$}: We will use a simpler variant of the $\PKPDIEF$ formulation introduced in the main paper as a proof of concept. Namely, we have, 
\begin{align}
    \mu_{\theta}(X_{t-1},U_{t-1},B) = \sum_{i=1}^{d} \sigma(\boldsymbol{\delta})_i \odot g_i(S_{t-1},U_{t-1},B),
    \label{eq:simplepkpd}
\end{align}
where each $\boldsymbol{\delta}$ is a learned vector of weights and $\sigma$ refers to a softmax on the weights. Note that the $\PKPDIEF$ introduced in the main paper is a generalization of Equation \ref{eq:simplepkpd}; the primary difference is that the attention mechanism allows the weights to be a function of the prior state, which enables the weights to change over time. 

\textit{Maximum Likelihood Estimation of $\theta$:} We learn the model by maximizing $\max_{\theta}\log p(\mathbf{X}|\mathbf{U},B)$. Using the factorization structure in the joint distribution of the generative model, we obtain: 
$\log p(\mathbf{X}|\mathbf{U},B) = \sum_{t=1}^T \log p(X_t|X_{t-1}, U_{t-1}, B)$. Each $\log p(X_t|X_{t-1}, U_{t-1}, B)$ is estimable as the log-likelihood of the observed multi-variate $X_t$
under a Gaussian distribution
whose (diagonal) variance is a function $\Sigma_{\theta}(X_{t-1},U_{t-1}, B)$
and whose mean is given by the transition function, $\mu_{\theta}(X_{t-1}, U_{t-1},B)$. Since each $\log p(X_t|X_{t-1}, U_{t-1}, B)$ is a differentiable function of $\theta$, its sum is differentiable as well, and we may use automatic differentiation to derive gradients of the log-likelihood with respect to $\theta$ in order to perform gradient ascent. When any dimension of $X_t$ is missing,
that dimension's log-likelihood is ignored (corresponding to marginalization over that random variable) during learning. 

\textit{Hyperparameters:} We present the results of the hyperparameter search on the datasets that we study. 
\begin{itemize}
    \item $\FOMMlinear$
    \begin{enumerate}
        \item \textit{Synthetic:} L$1$ regularization on all parameters with strength $0.1$
        \item \textit{ML-MMRF:} L$1$ regularization on all parameters with strength $0.1$
    \end{enumerate}
    \item $\FOMMnl$
    \begin{enumerate}
        \item \textit{Synthetic:} Hidden layer dimension $200$, L$1$ regularization on all parameters with strength $0.1$
        \item \textit{ML-MMRF:} Hidden layer dimension $300$, L$1$ regularization on all parameters with strength $0.1$
    \end{enumerate}
    \item $\FOMMpkpd$
    \begin{enumerate}
        \item \textit{Synthetic:} L$1$ regularization on subset of parameters with strength $0.1$
        \item \textit{ML-MMRF:} L$1$ regularization on subset of parameters with strength $0.1$
    \end{enumerate}
\end{itemize}

\paragraph{Gated Recurrent Neural Network (GRUs):} GRUs \citep{cho2014properties} are auto-regressive models of sequential observations i.e.
$p(\mathbf{X}|\mathbf{U},B) = \prod_{t=1}^{T}p(X_t|X_{<t}, U_{<t}, B)$). GRUs use an intermediate hidden state $h_t\in\R{H}$ at each time-step as a proxy for what the model has inferred about the sequence of data until $t$. The GRU dynamics govern how $h_{t}$ evolves via
an update gate $F_t$, and a reset gate $R_t$:
\begin{align}
\textstyle
    &F_t = \sigma(W_z\cdot [X_t,U_t,B] + V_z h_{t-1} + b_z), \\
    &R_t = \sigma(W_r\cdot [X_t,U_t,B] + V_r h_{t-1} + b_r) \nonumber\\
    &h_t = F_t \odot h_{t-1} + (1-F_t) \odot \nonumber\\
    &\tanh(W_h\cdot [X_t,U_t,B] + V_h(R_t \odot h_{t-1}) + b_h) \nonumber%
\end{align}
$\theta=\{$ $W_z,W_r,W_h\in\R{H \times (M+L+J)};V_z,V_r,V_h\in\R{H \times H}; b_z,b_r,b_h\in\R{H}\}$ are learned parameters and
$\sigma$ is the sigmoid function. 
The effect of interventions may be felt in any of the above time-varying representations and so the "transition function" in the GRU is distributed across the computation of the forget gate, reset gate and the hidden state, i.e. $S_t=[F_t,R_t,h_t]$. We refer to this model as $\RNN$.

\textit{$\PKPDIEF$ for $\RNNpkpd$}: We take the output of Equation \ref{eq:simplepkpd}, $o_t = \mu_{\theta}(X_{t-1}, U_{t-1}, B)$, and divide it into three equally sized vectors: $o_t^f,o_t^r,o_t^h$. Then,
  \begin{align}
      F_t &= \sigma(o_{t}^f + V_z h_{t-1} + b_z)\nonumber\\
      R_t &= \sigma(o_{t}^r + V_r h_{t-1} + b_r) \nonumber\\
      h_t &= F_t \odot h_{t-1} \nonumber\\ 
      &+(1-F_t) \odot \tanh(o_{t}^h + V_h(R_t \odot h_{t-1}) + b_h) \nonumber
  \end{align}

\textit{Maximum Likelihood Estimation of $\theta$:} We learn the model by maximizing $\max_{\theta}\log p(\mathbf{X}|\mathbf{U},B)$. Using the factorization structure in the joint distribution of the generative model, we obtain: 
$\log p(\mathbf{X}|\mathbf{U},B) = \sum_{t=1}^T \log p(X_t|X_{<t}, U_{<t}, B)$. 
At each point in time the hidden state of the GRU, $h_t$, 
summarizes $X_{<t}, U_{<t}, B$. Thus, the model assumes $X_t\sim \mathcal{N}(\mu_{\theta}(h_t),\Sigma_{\theta}(h_t))$.

At each point in time, $\log p(X_t|X_{<t}, U_{<t}, B)$ is the log-likelihood of a multi-variate Gaussian
distribution which depends on $\theta$. As before, we use automatic differentiation to derive gradients of the log-likelihood with respect to $\theta$ in order to perform gradient ascent. When any dimension of $X_t$ is missing, that dimension's log-likelihood is ignored (corresponding to marginalization over that random variable) during learning. 

\textit{Hyperparameters:} We present the results of the hyperameter search on the datasets that we study. 
\begin{itemize}
    \item $\RNN$
    \begin{enumerate}
        \item \textit{Synthetic:} Hidden layer dimension 500, L$2$ regularization on all parameters with strength $0.1$
        \item \textit{ML-MMRF:} Hidden layer dimension 250, L$2$ regularization on all parameters with strength $0.1$
    \end{enumerate}
    \item $\RNNpkpd$
    \begin{enumerate}
        \item \textit{Synthetic:} Hidden layer dimension 500, L$2$ regularization on subset of parameters with strength $0.01$
        \item \textit{ML-MMRF:} Hidden layer dimension 500, L$2$ regularization on subset of parameters with strength $0.01$
    \end{enumerate}
\end{itemize}

\newpage 

\section{Synthetic Dataset\label{sec:app_syn_data}}
\begin{figure*}[h!]
    \centering
    \includegraphics[width=\textwidth]{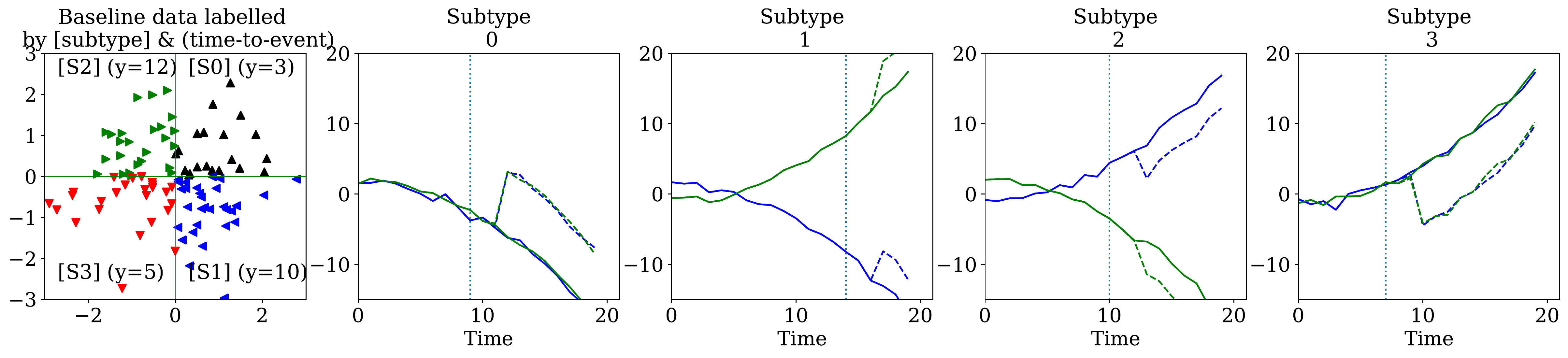}
    \caption{\small\textbf{Visualization of synthetic data:} Left: A visualization of "patient"'s baseline data (colored and marked by patient subtype). Right four plots: Examples of patient's longitudinal trajectories along with treatment response. The blue and green longitudinal data denote two different patient biomarkers. The solid blue and green lines are what the trajectories would be with no intervention whereas the dotted blue and green lines are what the trajectories are with an intervention. Gray-dotted line represents an intervention. The subtypes may, optionally, be correlated with patient outcomes as highlighted using the values of $y$.}
    \label{fig:syn_data}
\end{figure*}

Below, we outline the general principles that the synthetic data we design is based on: 

\begin{itemize}
    \item We sample six random baseline values from a standard normal distribution. 
    \item Two of the six baseline values determine the natural (untreated) progression of the  two-dimensional longitudinal trajectories. They do so as follows: depending on which quadrant the baseline data lie in, we assume that the patient has one of four subtypes. 
    \item Each of the four subtypes typifies different patterns by which the biomarkers behave such as whether they both go up, both go down, one goes up, one goes down etc. To see a visual example of this, we refer the reader to Figure \ref{fig:syn_data} (left).
\end{itemize}

\textbf{Baseline} The generative process for the baseline covariates is $B\sim\mathcal{N}(0;\mathbb{I});B\in\R{6}$.

\textbf{Treatments (Interventions):} 
There is a single drug (denoted by a binary random variable) that may be withheld (in the first line of therapy) or prescribed in the second line of therapy. 
For each patient $d_i\sim\text{Unif.}[0,18]$ denotes when the single drug is administered (and the second line of treatment begins). $d_i$ is the point at which the local clock resets. 
We can summarize the generative process for the treatments as follows: 
\begin{align}
&d\sim \text{Unif.}[0,18] \nonumber \\ 
&U_t = 0 \text{ if } t<d \text{ and } 1 \text{ otherwise}\nonumber \\
&\text{line}_t[0] = 1 \text{ if } t<d \text{ and } 0 \text{ otherwise}\nonumber \\
&\text{line}_t[1] = 0 \text{ if } t<d \text{ and } 1 \text{ otherwise}
\end{align}
where $\text{line}_t[0],\text{line}_t[1]$ denote the one-hot encoding for line of therapy. Next, we use the Neural Treatment Exponential (Equation \ref{eq:trt_exp_main} in the main paper) to generate a treatment response in the data. 
The functional form of the Treatment Exponential function (TE) is re-stated below for convenience, %
\begin{align}
    \text{TE}(\text{lc}_t) = \begin{cases}
    b_0 + \alpha_{1} / [1+\exp (-\alpha_{2}(\text{lc}_t-\frac{\gamma_{l}}{2}))], \\\qquad\text{if } 0 \leq \text{lc}_t < \gamma_l \\
    b_l + \alpha_{0} / [1+\exp (\alpha_{3}(\text{lc}_t-\frac{3\gamma_{l}}{2}))], \\ \qquad \text{if } \text{lc}_t \geq \gamma_l
  \end{cases}
\end{align}
The parameters that we use to generate the data are: $\alpha_2 = 0.6, \alpha_3 = 0.6, \gamma_l = 2, b_l = 3,$ and $\alpha_1 = [10,5,-5,-10]$, which we vary based on patient subtype. We set $\alpha_0 = (\alpha_1 + 2b_0 - b_l) / (1 + \exp(-\alpha_3 \gamma_l)/2)$ to ensure that the treatment effect peaks at $t = \text{lc}_t  + \gamma_l$ and $b_0 = -\alpha_1 / (1 + \exp(\alpha_2 \cdot \gamma_l / 2))$ for attaining $\text{TE}(0) = 0$.

\textbf{Biomarkers:} We are now ready to describe the full generative process of the longitudinal biomarkers. 

\begin{align}
\label{eqn:synthetic_gen}
    &B_{1\ldots,6}\sim \mathcal{N}(0;I), \nonumber \\
    &f_d(t) = 2 - 0.05t -0.005t^2,\\ 
    &f_u(t) =-1+0.0001t+0.005t^2, \nonumber \\
    &X_1(t); X_2(t) =  \\
    &\begin{cases} 
      f_{d}(t)+\text{TE}(\text{lc}_t)+\mathcal{N}(0, 0.25); \; f_{d}(t) + \text{TE}(\text{lc}_t) \nonumber \\ \quad \quad + \mathcal{N}(0, 0.25), B_1\geq 0, B_2\geq 0 \text{ if subtype } 1\\
      f_{d}(t)+\text{TE}(\text{lc}_t)+ \mathcal{N}(0, 0.25); \; f_u(t) + \text{TE}(\text{lc}_t) \nonumber \\ \quad \quad + \mathcal{N}(0, 0.25), B_1\geq 0, B_2< 0 \text{ if subtype } 2\\
      f_u(t) + \text{TE}(\text{lc}_t) + \mathcal{N}(0, 0.25); \; f_{d}(t) + \text{TE}(\text{lc}_t) \nonumber  \\ \quad \quad + \mathcal{N}(0, 0.25), B_1< 0, B_2\geq 0 \text{ if subtype } 3\\
      f_u(t) + \text{TE}(\text{lc}_t) + \mathcal{N}(0, 0.25); \; f_u(t) + \text{TE}(\text{lc}_t) \nonumber \\ \quad \quad + \mathcal{N}(0, 0.25), B_1< 0, B_2< 0 \text{ if subtype } 4,
   \end{cases}\nonumber 
\end{align}

Intuitively, the above generative process captures the idea that without any effect of treatment, the biomarkers follow the patterns implied by the subtype (encoded in the first two dimensions of the baseline data). However the effect of interventions is felt more prominently after
the $d$, the random variable denoting time at which treatment was prescribed. 

\section{The Multiple Myeloma Research Foundation CoMMpass Study\label{sec:mm_data_app}}
Here, we elaborate upon the data made available by the Multiple Myeloma Research Foundation in the IA13 release of data. We will make code available to go from the files released by the MMRF study to numpy tensors that may be used in any machine learning framework.

\textbf{Inclusion Criteria:} To enroll in the CoMMpass study, patients must be newly diagnosed with symptomatic multiple myeloma, which coincides with the start of treatment. Patients must be eligible for treatment with an immunomodulator or a proteasome inhibitor, two of the most common first line drugs, and they must begin treatment within 30 days of the baseline bone marrow evaluation \cite{usrelating}.

\subsection{Features}
\label{sec:mm-features}

\textbf{Genomic Data:} RNA-sequencing of CD38+ bone marrow cells was available for 769 patients. Samples were collected at initiation into the study, pre-treatment. For these patients, we used the Seurat package version 2.3.4 \cite{butler2018integrating} in R to identify variable genes, and we then limit downstream analyses to these genes. We use principal component analysis (PCA) to further reduce the dimensionality of the data. The projection of each patient's gene expression on to the first 40 principal components serves as the genetic features used in our model.

\textbf{Baseline Data:} Baseline data includes PCA scores, lab values at the patient's first visit, gender, age, and the revised ISS stage. The baseline data also includes binary variables detailing the patient's myeloma subtype, including whether or not they have heavy chain myeloma, are IgG type, IgA type, IgM type, kappa type, or lambda type.
Additionally, several labs are measured at baseline, as well as longitudinally at subsequent visits. We detail these labs in the next sub-section.
The genetic and baseline data jointly comprise $B$.

\textbf{Longitudinal Data:} Longitudinal data is measured approximately every 2 months and includes lab values and treatment information. 

The biomarkers are real-valued numbers whose values evolve over time. They include: 
absolute neutrophil count (x$10^{9}$/l), albumin (g/l), blood urea nitrogen (mmol/l), calcium (mmol/l), serum creatinine (umol/l), glucose (mmol/l), hemoglobin (mmol/l), serum kappa (mg/dl),
serum m protein (g/dl),  platelet count x$10^9$/l, total protein (g/dl), white blood count  x$10^9$/l,  serum IgA (g/l), serum IgG (g/l), serum IgM (g/l), serum lambda (mg/dl).

Treatment information includes the line of therapy (we group all lines beyond line $3$ as line $3+$) the patient is on at a given point in time, and the local clock denoting the time elapsed since the last line of therapy. We also include the following treatments as (binary, indicating prescription) features in our model: lenalidomide, dexamethasone, cyclophosphamide, carfilzomib, bortezomib. The aforementioned are the top five drugs by frequency in the MMRF dataset. This dataset has significant missingness, with $\sim 66\%$ of the longitudinal markers missing. In addition, there is right censorship in the dataset, with around $25\%$ of patients getting censored over time.

\subsection{Data Processing}
Longitudinal biomarkers $\mathbf{X}$: Labs are first clipped to five times the median value to correct for outliers or data errors in the registry. They are then normalized to their healthy ranges (obtained via a literature search) as (unnormalized labs - healthy maximum value), and then multiplied by a lab-dependent scaling factor to ensure that most values lie within the range $[-8,8]$. Missing values are represented as zeros, but a separate mask tensor, where $1$ denotes observed and $0$ denotes
missing, is used to marginalize out missing variables during learning.

Baseline $B$: The biomarkers in the baseline are clipped to five times their median values. Patients without gene expression data (in the PCA features) are assigned the average normalized PCA score of their five nearest neighbors, using the Minkowski distance metric calculated on FISH features, ISS stage, and age. 

\newpage 
\section{FOMM and GRU Experiments}
\begin{table*}
\centering
\begin{tabular}[t]{c c c c c c c c c}
\toprule
 Dataset & \textbf{\begin{tabular}[c]{@{}c@{}}Held-out Neg\\ Log Likelihood\end{tabular}} &    \textbf{\begin{tabular}[c]{@{}c@{}}FOMM\\ Linear\end{tabular}} &  \textbf{\begin{tabular}[c]{@{}c@{}}FOMM\\ Nonlinear\end{tabular}} &
 \textbf{\begin{tabular}[c]{@{}c@{}}FOMM\\ PK-PD\end{tabular}} & \textbf{\begin{tabular}[c]{@{}c@{}}FOMM \\ MOE\end{tabular}} &  \textbf{\begin{tabular}[c]{@{}c@{}}GRU\end{tabular}} & \textbf{\begin{tabular}[c]{@{}c@{}}GRU\\ PK-PD\end{tabular}} & 
 \begin{tabular}[c]{@{}c@{}}\textbf{SSM}\\ \textbf{PK-PD} \\ (NELBO)\end{tabular}\\
\midrule                                                    
  \begin{tabular}[c]{@{}c@{}}ML-MMRF\end{tabular} &   &      92.80 &      97.53 & \textbf{90.26} & 97.26 & \textbf{89.89} & 99.98 & \textbf{61.54}\\ 
\bottomrule
\end{tabular}
\bigbreak
\scalebox{0.9}{
\begin{tabular}[t]{c c c c c c}
    \toprule
     \textbf{Dataset} &   \textbf{\begin{tabular}[c]{@{}c@{}}FOMM\\ PK-PD\end{tabular}} vs.\textbf{ \begin{tabular}[c]{@{}c@{}}FOMM\\ Linear\end{tabular}} &  \textbf{\begin{tabular}[c]{@{}c@{}}FOMM\\ PK-PD\end{tabular}} vs.\textbf{\begin{tabular}[c]{@{}c@{}}FOMM\\ NL\end{tabular}} &
     \textbf{\begin{tabular}[c]{@{}c@{}}FOMM\\ PK-PD\end{tabular}} vs. \textbf{\begin{tabular}[c]{@{}c@{}}FOMM\\ MOE\end{tabular}} & 
     \textbf{\begin{tabular}[c]{@{}c@{}}GRU\\ PK-PD\end{tabular}} vs. \textbf{\begin{tabular}[c]{@{}c@{}}GRU\end{tabular}} & ----- \\
    \midrule                                                    
      \begin{tabular}[c]{@{}c@{}}ML-MMRF\end{tabular} &  0.792 (0.405)    &    0.668 (0.457)   & 0.510 (0.490) & 0.406 (0.489) & -----\\ 
     \midrule 
    \begin{tabular}[c]{@{}c@{}}\end{tabular} &
     \textbf{\begin{tabular}[c]{@{}c@{}}FOMM\\ Linear\end{tabular}} & 
     \textbf{\begin{tabular}[c]{@{}c@{}}FOMM\\ NL\end{tabular}} &
 \textbf{\begin{tabular}[c]{@{}c@{}}FOMM\\ PK-PD\end{tabular}} & \textbf{\begin{tabular}[c]{@{}c@{}}GRU\end{tabular}} & 
 \textbf{\begin{tabular}[c]{@{}c@{}}GRU\\ PK-PD\end{tabular}}\\
 \midrule
\begin{tabular}[c]{@{}c@{}}Synthetic (50 samples)\end{tabular} & 71.06 +/- .03 & 58.80 +/- .03 & \textbf{56.81 +/- .04} & 56.65 +/- .11 & \textbf{53.49 +/- .04} \\
 \midrule 
 Synthetic (1000 samples) & 62.93 +/- .03 & \textbf{57.16 +/- .03} & 57.81 +/- .02 & 31.09 +/- .02 & \textbf{29.27 +/- .01} \\ 
    \bottomrule
\end{tabular}
}
\caption{\small \textbf{Generalization of FOMM and GRU models on ML-MMRF and Synthetic Data:}
\textit{Top}: Lower is better. We report negative log-likelihood averaged across five folds of held-out data. \textit{Bottom}: \textit{ML-MMRF}: We report pairwise comparisons of models trained on ML-MMRF. Higher is better. Each number is the fraction (with std. dev.) of held-out patients for which the model that uses $\PKPDIEF$ has a lower negative log-likelihood than a model in the same family that uses a different transition function. \textit{Synthetic}: Lower is better. We report held-out negative log likelihood with std. dev. on FOMM and GRU to study generalization in the synthetic setting.}
\label{tab:tabglobapp1p}
\end{table*}

\textit{$\PKPDIEF$ improves generalization in both \textbf{FOMM} and \textbf{GRU} models on synthetic data:} Table \ref{tab:tabglobapp1p} (bottom) depicts negative log-likelihoods on held-out synthetic data 
across different models, where a lower number implies better generalization. 
The non-linearity of the synthetic data makes unsupervised learning a challenge for $\FOMMlinear$ at $50$ samples, allowing $\FOMMpkpd$ to easily outperform it. In contrast, $\FOMMnl$ can capture non-linearities in the data, making it a strong baseline even at $50$ samples. Yet, $\FOMMpkpd$ outperforms it, emphasizing the utility of $\PKPDIEF$ in when data is scarce. 
At $1000$ samples, $\FOMMnl$ is able to learn enough about the dynamics to improve its performance relative to $\FOMMpkpd$. $\RNN$ is a strong model on this dataset, but at both sample sizes, the $\RNNpkpd$ improves generalization. 

\textit{\textbf{FOMM} and \textbf{GRU} generalization performance on ML-MMRF :} We observe improvements in generalization across FOMMs with the use of $\PKPDIEF$. However, we do not see discernible gains from $\RNNpkpd$, perhaps due to missingness in the data, which also results in the GRUs generalizing worse than SSM models based on negative log likelihood (Table \ref{tab:tabglobapp1p} (top)). Overall, the GRU, due to the distributed nature of its "transition function", is a harder model to use $\PKPDIEF$ in, although figuring how to do so effectively is a valuable direction for future work.

\newpage 
\section{Experiments on Semi-synthetic Dataset}
In this section, we cover how to generate the semi-synthetic dataset. We then provide experimental results on generalization performance as well as a result in a model misspecification scenario.

\textbf{Semi-synthetic data:} We train the $\SSMpkpd$ model on the ML-MMRF dataset and generate samples
from the model. For each sequence of treatments, we generate $30$ random samples per training data point resulting in a dataset of size $14000$. Then we uniformly at random sample $1000$ samples from that pool to form our training set. We perform a similar procedure to generate several held-out sets (size $87000$ samples each). This semi-synthetic dataset allows us to 
ask questions about generalization on data with statistics similar to ML-MMRF.

\textbf{Generalization and Model Misspecification} \textit{$\SSMpkpd$ generalizes well with fewer samples:} At 1000 samples, we find that the $\SSMpkpd$ models generalize better than the baselines, where a lower, more negative number implies better generalization. We see that with few samples, $\SSMnl$ and $\SSMmoe$ overfit. However, when sharply increasing the number of samples to 20000, both models recover their performance and even begin to outperform $\SSMpkpd$. This result further solidifies the generalization capability of our proposed $\SSMpkpd$ model in a data-scarce setting as well as the difficulty of learning a nonlinear model that does not overfit.

\textit{$\SSMpkpd$ continues to generalize well even when it is mis-specified:} We run a similar experiment to what we ran on ML-MMRF, where we take out the Neural Treatment Exponential mechanism function from $\mu_{\theta}$ and instead opt for using a linear function. We see that $\SSMpkpd$ \textbf{w/o TExp} performs comparably to a $\SSMpkpd$ with the Neural Treatment Exponential mechanism, providing further evidence that our architecture can recover the unknown intervention effect in the data via a combination of related mechanism functions. 
\begin{table}[h!]
\centering
\caption{\small \textbf{Held-out NELBO on semi-synthetic data}: Trained on 1K or 20K samples and evaluated on ~87K samples, we show the mean test NELBO and standard deviation across five test sets. 
}
\scalebox{0.7}{
\begin{tabular}[t]{||c c c c||}
\toprule
    Sample Size &
 \textbf{\begin{tabular}[c]{@{}c@{}}SSM Linear \end{tabular}} &
 \textbf{\begin{tabular}[c]{@{}c@{}}SSM NL\end{tabular}} & 
 \textbf{\begin{tabular}[c]{@{}c@{}}SSM MOE \end{tabular}}\\
 \midrule
 1000 & -211.54 +/- 49.61 & -192.86 +/- 25.77 & -266.37 +/- 10.42 \\ 
 \midrule 
 & 
  \textbf{\begin{tabular}[c]{@{}c@{}}SSM PK-PD \end{tabular}} & \textbf{\begin{tabular}[c]{@{}c@{}}SSM PK-PD \\ (w/o TExp) \end{tabular}} & ------\\
  \midrule 
  1000 &  \textbf{-294.00 +/- 10.25} & \textbf{-295.44 +/- 8.18}  &------\\
  \midrule 
     &
 \textbf{\begin{tabular}[c]{@{}c@{}}SSM Linear \end{tabular}} &
 \textbf{\begin{tabular}[c]{@{}c@{}}SSM NL\end{tabular}} & 
 \textbf{\begin{tabular}[c]{@{}c@{}}SSM MOE \end{tabular}}\\
 \midrule
 20000 & -322.36 +/- 0.30 & -316.09 +/- 0.18 &  \textbf{-322.22 +/- 0.22} \\ 
 \midrule 
 & 
  \textbf{\begin{tabular}[c]{@{}c@{}}SSM PK-PD \end{tabular}} & \textbf{\begin{tabular}[c]{@{}c@{}}SSM PK-PD \\ (w/o TExp) \end{tabular}} & ------\\
  \midrule 
  20000 & -319.57 +/- 0.23 & -315.60 +/- 0.40 & ------\\
\bottomrule
\end{tabular}
}
\label{tab:tabglob_semi}
\end{table}

\textbf{Hyperparameters}: We present the best hyperparameters for each model at each sample size. We search over the ranges as described in the main paper; however, at 20000 samples, we train for 1000 epochs instead of 15000 epochs, which we found to be a more stable training configuration.
\begin{itemize}
    \item $\SSMlinear$
    \begin{enumerate}
        \item 1000 samples: State space dimension 64, L2 regularization on all parameters with strength 0.01
        \item 20000 samples: State space dimension 128, L2 regularization on all parameters with strength 0.01
    \end{enumerate}
    \item $\SSMnl$
    \begin{enumerate}
        \item 1000 samples: State space dimension 48, hidden layer dimension on neural network: 300, L2 regularization on all parameters with strength 0.1
        \item 20000 samples: State space dimension 128, hidden layer dimension on neural network: 300, L2 regularization on all parameters with strength 0.01
    \end{enumerate}
    \item $\SSMmoe$
    \begin{enumerate}
        \item 1000 samples: State space dimension 48, hidden layer dimension on each MLP expert: 300, L2 regularization on all parameters with strength 0.01
        \item 20000 samples: State space dimension 128, hidden layer dimension on each MLP expert: 300, L2 regularization on all parameters with strength 0.01
    \end{enumerate}
    \item $\SSMpkpd$
    \begin{enumerate}
        \item 1000 samples: State space dimension 64, L2 regularization on subset of parameters with strength 0.01
        \item 20000 samples: State space dimension 128, L2 regularization on subset of parameters with strength 0.01
    \end{enumerate}
\end{itemize}

\newpage 
\section{Additional Analyses \label{ref:app_analyses}}

This section presents experimental results that provide an additional qualitative lens onto the $\PKPDIEF$.

\subsection{Exploring different strategies of sampling patient data using $\SSMpkpd$ on ML-MMRF \label{sec:C2}}

\begin{figure*}
\begin{subfigure}{\textwidth}
  \centering
  \includegraphics[width=\linewidth]{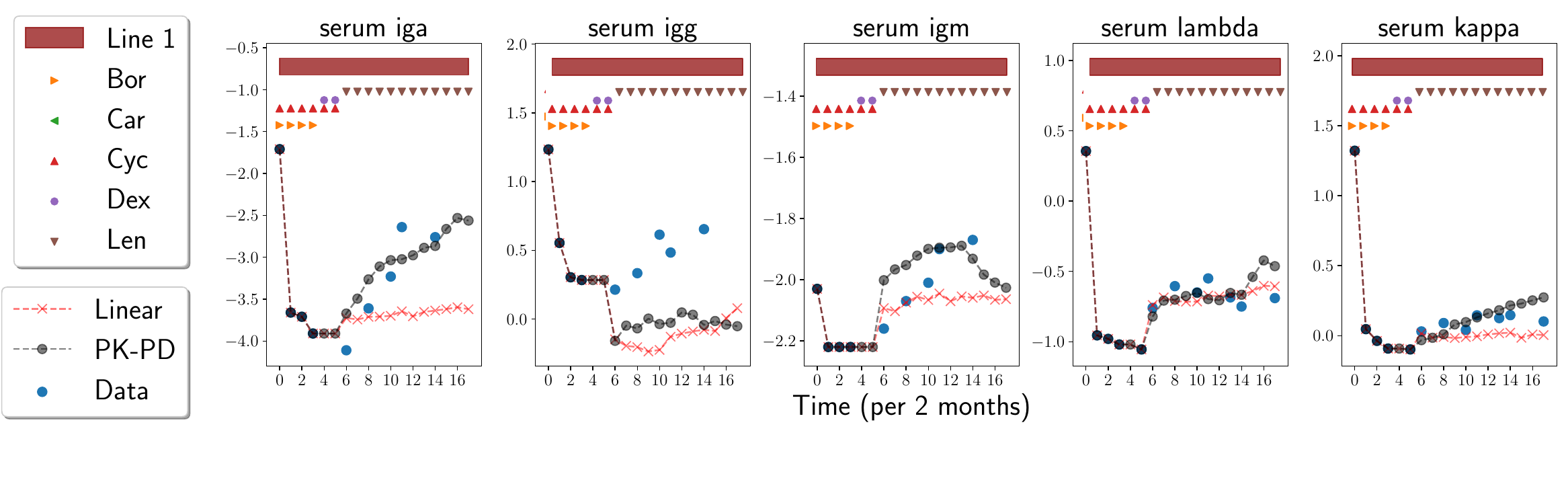}
  \caption{}
\end{subfigure}
\begin{subfigure}{\textwidth}
  \centering
  \includegraphics[width=\textwidth]{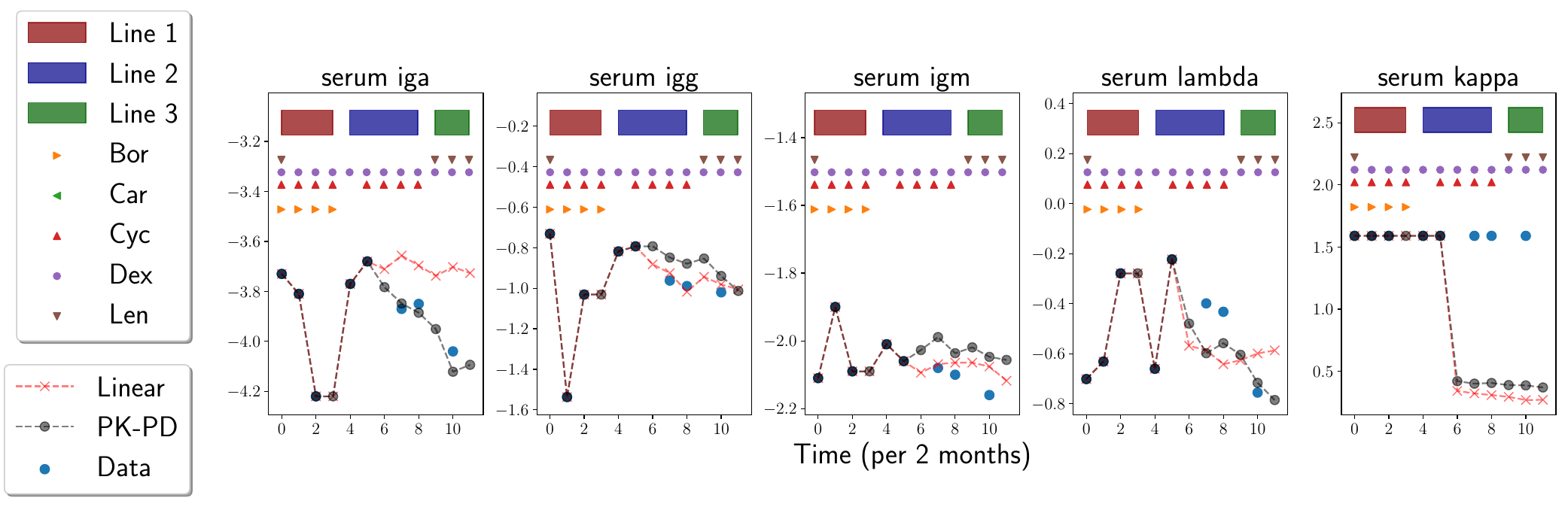}
  \caption{}
\end{subfigure}
\begin{subfigure}{\textwidth}
  \centering
  \includegraphics[width=\textwidth]{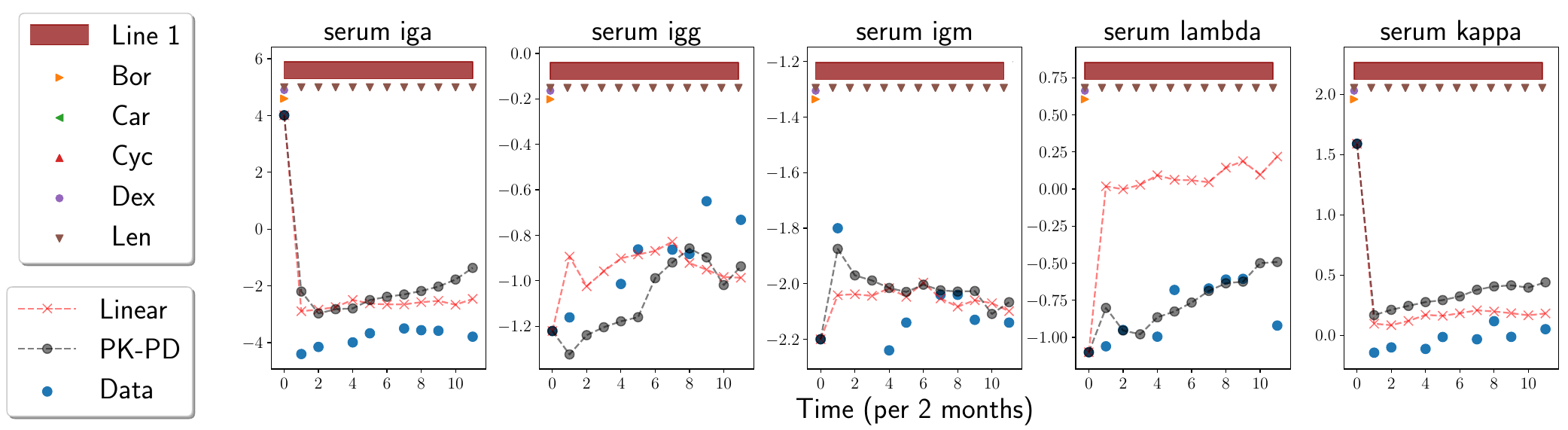}
  \caption{}
\end{subfigure}
\caption{\small\textbf{Forward samples from learned SSM models with differing conditioning strategies}: We visualize samples from $\textbf{SSM}_\text{PK-PD}$ (\textcolor{black}{o}) and $\textbf{SSM}_\text{linear}$ (\textcolor{red}{x}). Each row corresponds to a single patient, whereas each column represents a different biomarker for that patient. \textbf{a)}: We condition on 6 months of patient data and forward sample 2 years. \textbf{b)}: We condition on 1 year of patient data and forward sample 1 year. \textbf{c)}: We condition on a patient's baseline data and forward sample 2 years. As in the main paper, blue circles denote ground truth, and the markers above the trajectories represent treatments prescribed across time.}
\label{fig:app_samples}
\end{figure*}

\begin{figure*}[ht]
    \centering
    \includegraphics[width=\textwidth]{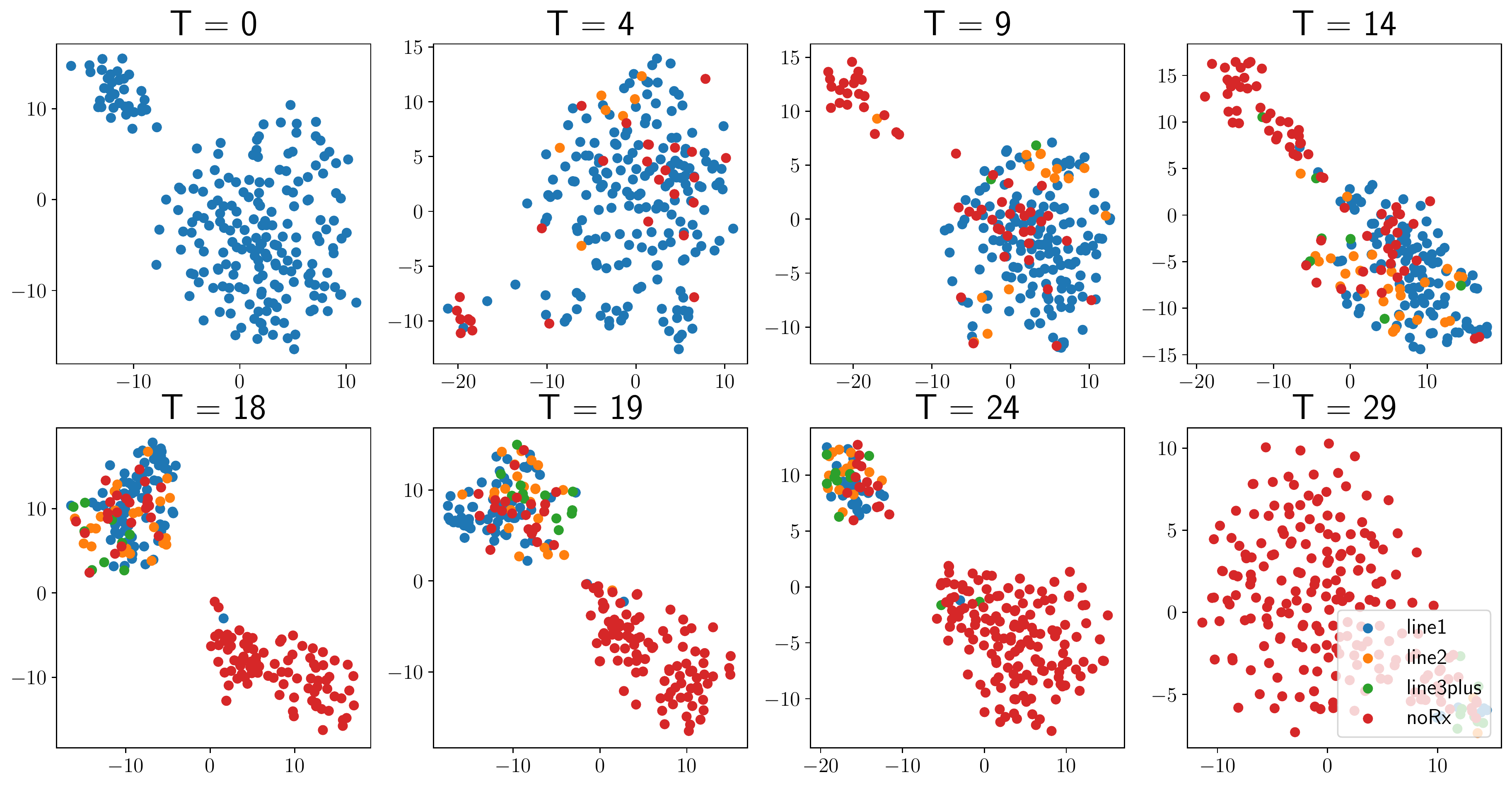}
    \caption{\small\textbf{$Z_{t}$ Visualizations:} We visualize the TSNE representations of each held-out patient's latent state, $Z_t$, over multiple time points, extending the analysis done in the main paper.
}
    \label{fig:zt_analysis}
\end{figure*}

In the main paper (Figure \ref{fig:mm_samples}), we show samples from SSM models trained on ML-MMRF, conditioned on a patient's first two years of data and the sequence of interventions they were prescribed. In each case, we additionally condition on the patient's baseline covariates.

Here, we experiment with different conditioning strategies. Let $C$ denote the point in time until which
we condition on patient data and $F$ denote the number of timesteps that we sample forward into the future. 
We limit our analysis to the subset of patients for which $C+F<=T$ where $T$ is the maximum number of time
steps for which we observe patient data. 

The samples we display are obtained as a consequence of averaging over three different samples, each of which
is generated (for the SSM) as follows: 
\begin{align}
    &Z\sim q_{\phi}(Z_C|Z_{C-1},X_{1:C},U_{0:C-1})\nonumber\\
    &Z_k \sim p_{\theta}(Z_k|Z_{k-1}, U_{k-1}, B) \quad k={C+1,\ldots,C+F}\nonumber\\
    &X_k \sim p_{\theta}(X_k|Z_k) \quad k={C+1,\ldots,C+F}\label{eqn:sup_sampling_procedure}
\end{align}

We study the following strategies for simulating patient data from the models. 
\begin{enumerate}
    \item Condition on 6 months of patient data, and then sample forward 2 years, 
    \item Condition on 1 year of a patient data and then sample forward 1 year,
    \item Condition on the baseline data of the patient and then sample forward 2 years.
\end{enumerate}

In Supp. Figure \ref{fig:app_samples}, we show additional samples from $\SSMpkpd$ when conditioning on differing amounts of data. 
Overall, in all three cases, $\SSMpkpd$ models capture treatment response better than one of the best performing baselines (i.e. $\SSMlinear$). 
For 1. (Figure \ref{fig:app_samples}a)), we see that $\SSMpkpd$ correctly captures that the serum IgA value goes up, while $\SSMlinear$ predicts that it will stay steady.
For 2. (Figure \ref{fig:app_samples}b)), $\SSMpkpd$ does well in modeling down-trends, as in serum IgA and serum lambda. 
For 3. (Figure \ref{fig:app_samples}c)), we similarly see that $\SSMpkpd$ captures the up-trending serum IgG and serum lambda.

\subsection{Analyzing the Latent State learned by $\SSMpkpd$ over Time}
In Supp Figure \ref{fig:zt_analysis}, we show the latent state of each held-out patient (reduced down to two dimensions via TSNE \citep{maaten2008visualizing}) over multiple time points, expanding on the two time points that were shown in Figure \ref{fig:lat_space_attention} of the main paper. As we saw before, early in the treatment course, the latent representations of the patients have no apparent structure. However, as time goes on, we find that the latent representations separate based on whether treatment is administered or not. 

\subsection{Deep Dive into $\SSMpkpd$ vs $\SSMlinear$ on ML-MMRF \label{sec:C4}}
\begin{figure*}[ht]
\centering
\begin{subfigure}{0.48\textwidth}
  \centering
  \includegraphics[width=\linewidth]{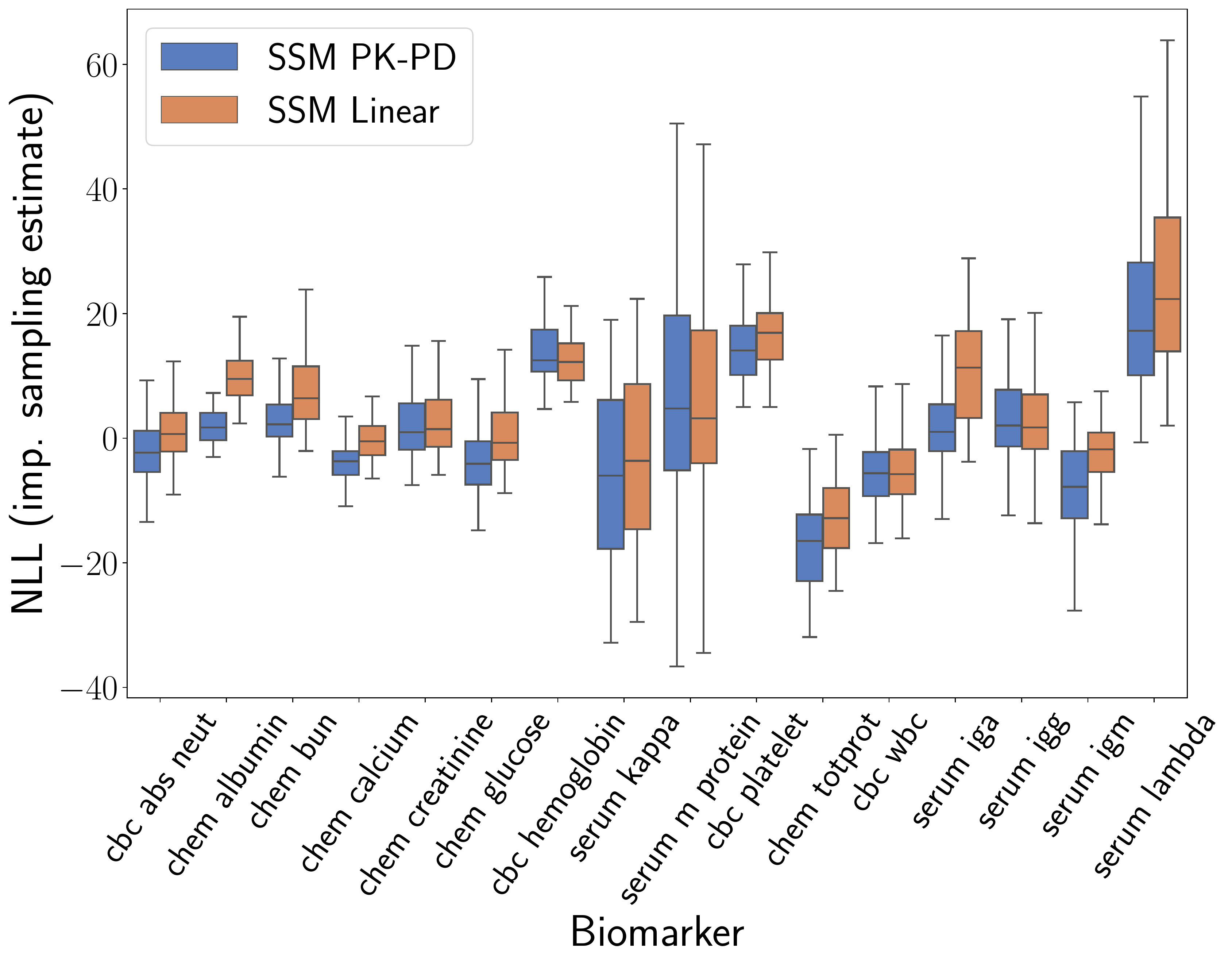}
  \caption{}
\end{subfigure}
\begin{subfigure}{0.48\textwidth}
  \centering
  \includegraphics[width=\linewidth]{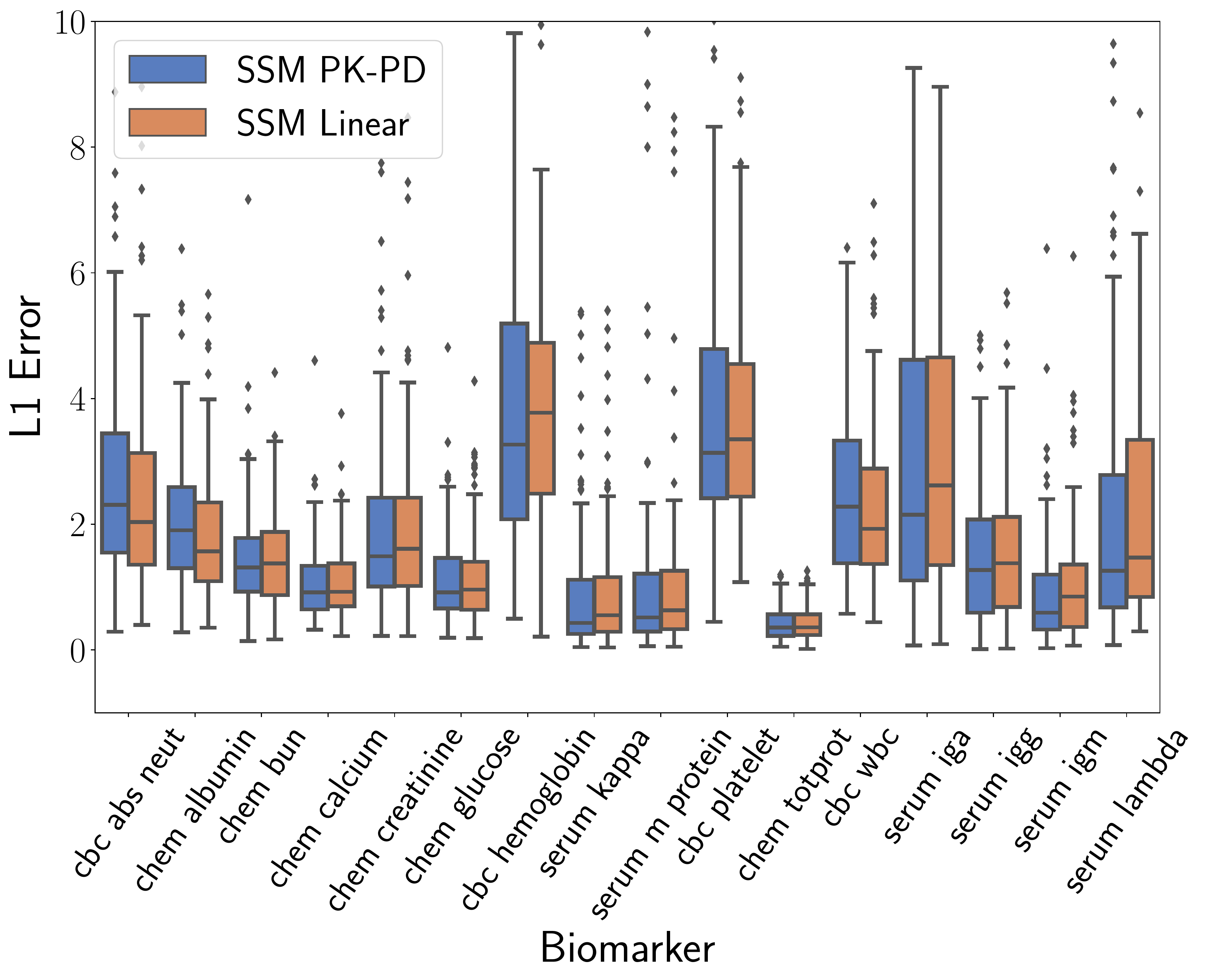}
  \caption{}
\end{subfigure}
\begin{subfigure}{0.48\textwidth}
  \centering
  \includegraphics[width=\linewidth]{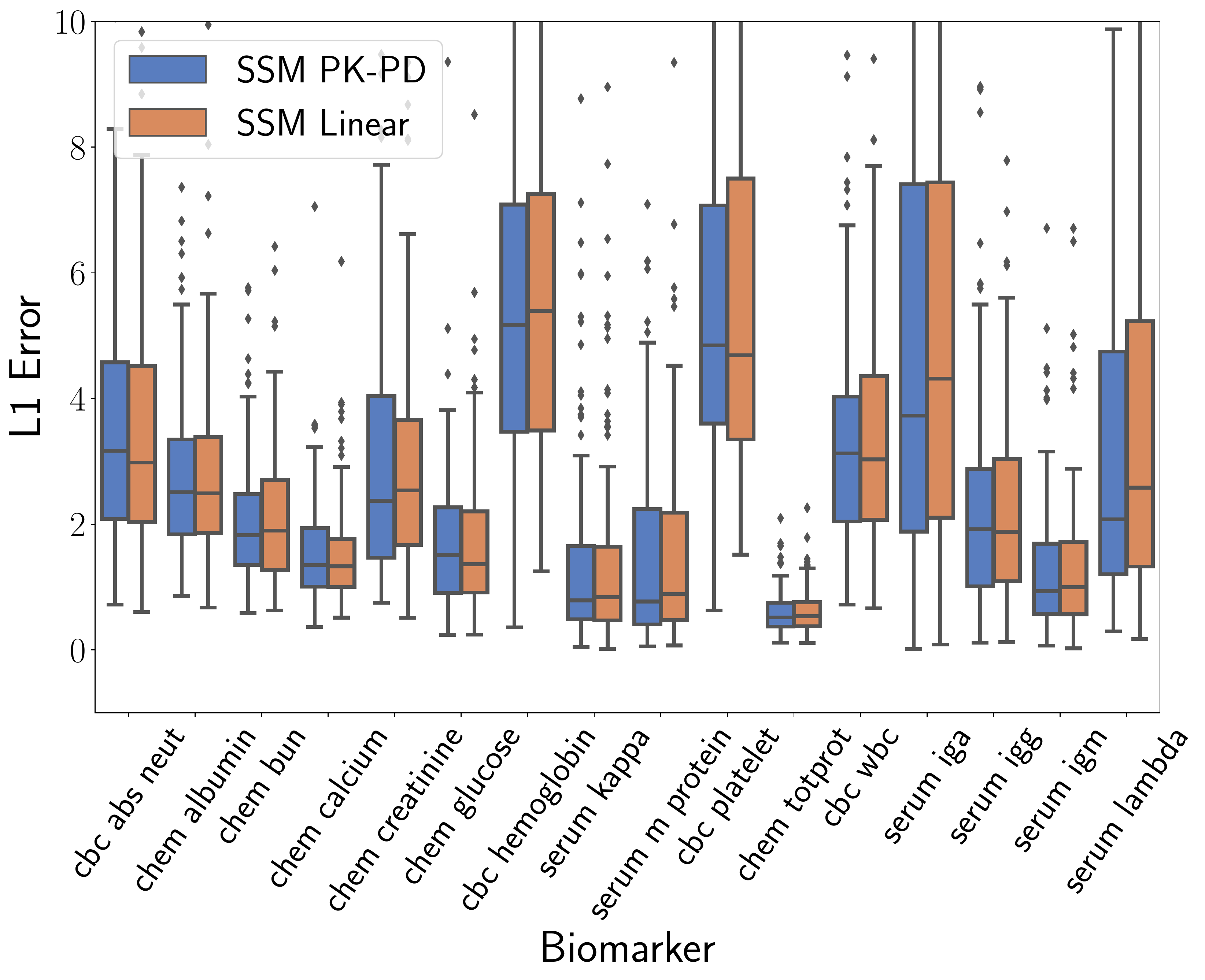}
  \caption{}
\end{subfigure}
\begin{subfigure}{0.48\textwidth}
  \centering
  \includegraphics[width=\linewidth]{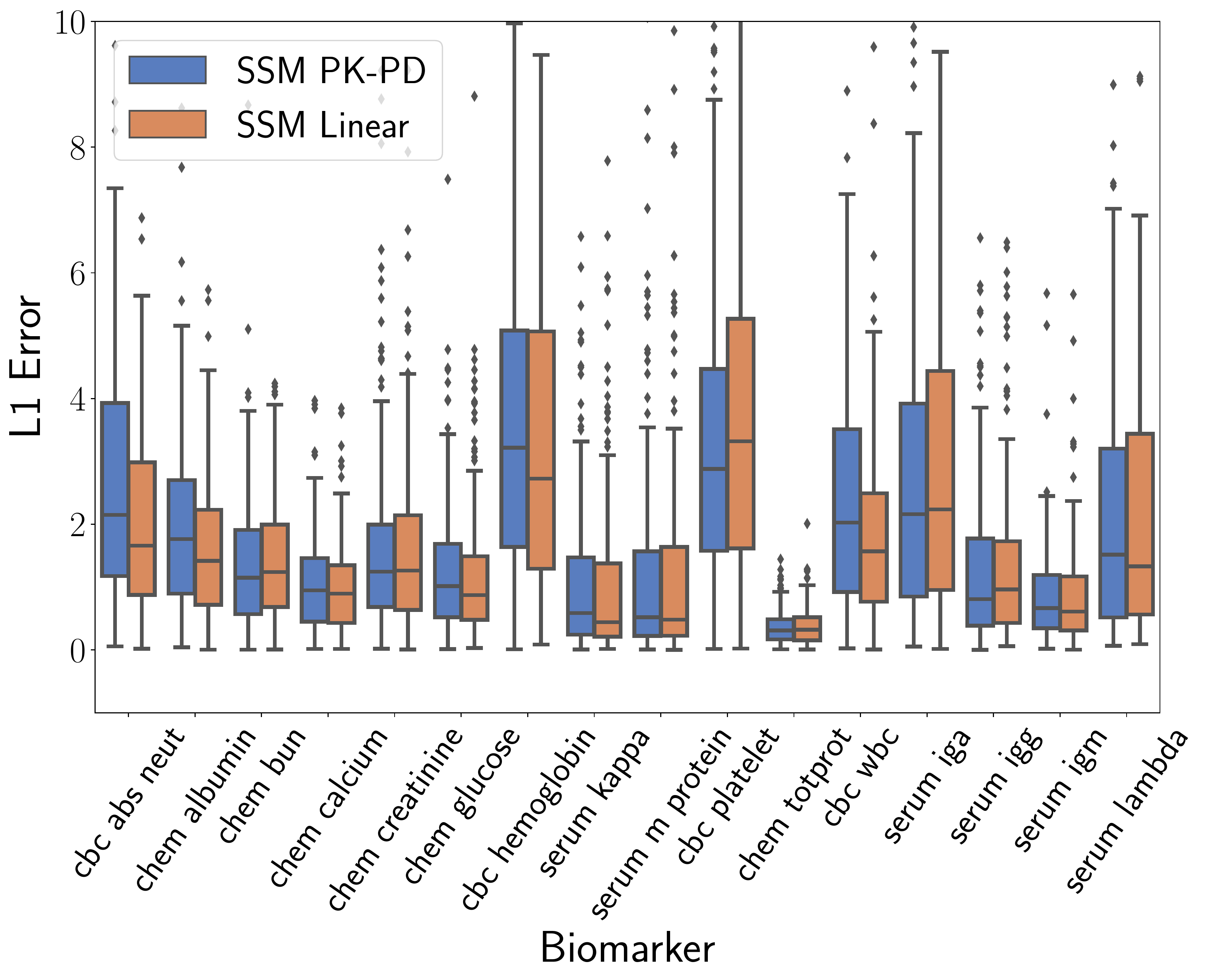}
  \caption{}
\end{subfigure}
\caption{\small\textbf{a) NLL estimates via Importance Sampling [top left]:} We estimate the NLL of $\SSMpkpd$ and $\SSMlinear$ for each feature, summed over all time points and averaged over all patients. \textbf{b) Condition on 6 months, forward sample 1 year [top right]:} We show L1 Prediction Error for forward samples over a 1 year time window conditioned on 6 months of patient data. At each time point, we compute the L1 error with the observed biomarker and sum these errors (excluding predictions for missing biomarker values) over the prediction window. We employ this procedure for each patient. \textbf{c) Condition on 6 months, forward sample 2 years [bottom left]:} We report L1 error for forward samples over a 2 year window conditioned on 6 months of patient data. \textbf{d) Condition on 2 years, forward sample 1 year [bottom right]:} Finally, we report L1 error for forward samples over a 1 year time window conditioned on 2 years of patient data.
}
\label{fig:bar_raw}
\end{figure*}

We are also interested in the absolute negative log likelihood measures and predictive capacity of the models at a per-feature level. In Supp. Figure \ref{fig:bar_raw}a), we use importance sampling to estimate the marginal negative log likelihood of $\SSMlinear$ and $\SSMpkpd$ for each covariate across all time points. Namely, we utilize the following estimator, 
\begin{align}
p(\mathbf{X}) \approx \frac{1}{S} \sum_{s=1}^{S} \frac{p(\mathbf{X}|\mathbf{Z}^{(s)})p(\mathbf{Z}^{(s)})}{q(\mathbf{Z}^{(s)}|\mathbf{X})},
\end{align}
akin to what is used in \cite{rezende2014stochastic}.
$\SSMpkpd$ has lower negative log likelihood compared to $\SSMlinear$ for several covariates, including neutrophil count, albumin, BUN, calcium, and serum IgA. This result is corroborated with the generated samples in Supp. Figure \ref{fig:app_samples}, which often show that the PK-PD model qualitatively does better at capturing IgA dynamics compared to the Linear model. In general, although there is a some overlap in the estimates of the likelihood under the two models for some features, it is reassuring to see that $\SSMpkpd$ does model the probability density of vital markers like serum IgA (which is often used by doctors to measure progression for specific kinds of patients), better than the baseline.

\begin{figure}
    \centering
    \includegraphics[scale=0.2]{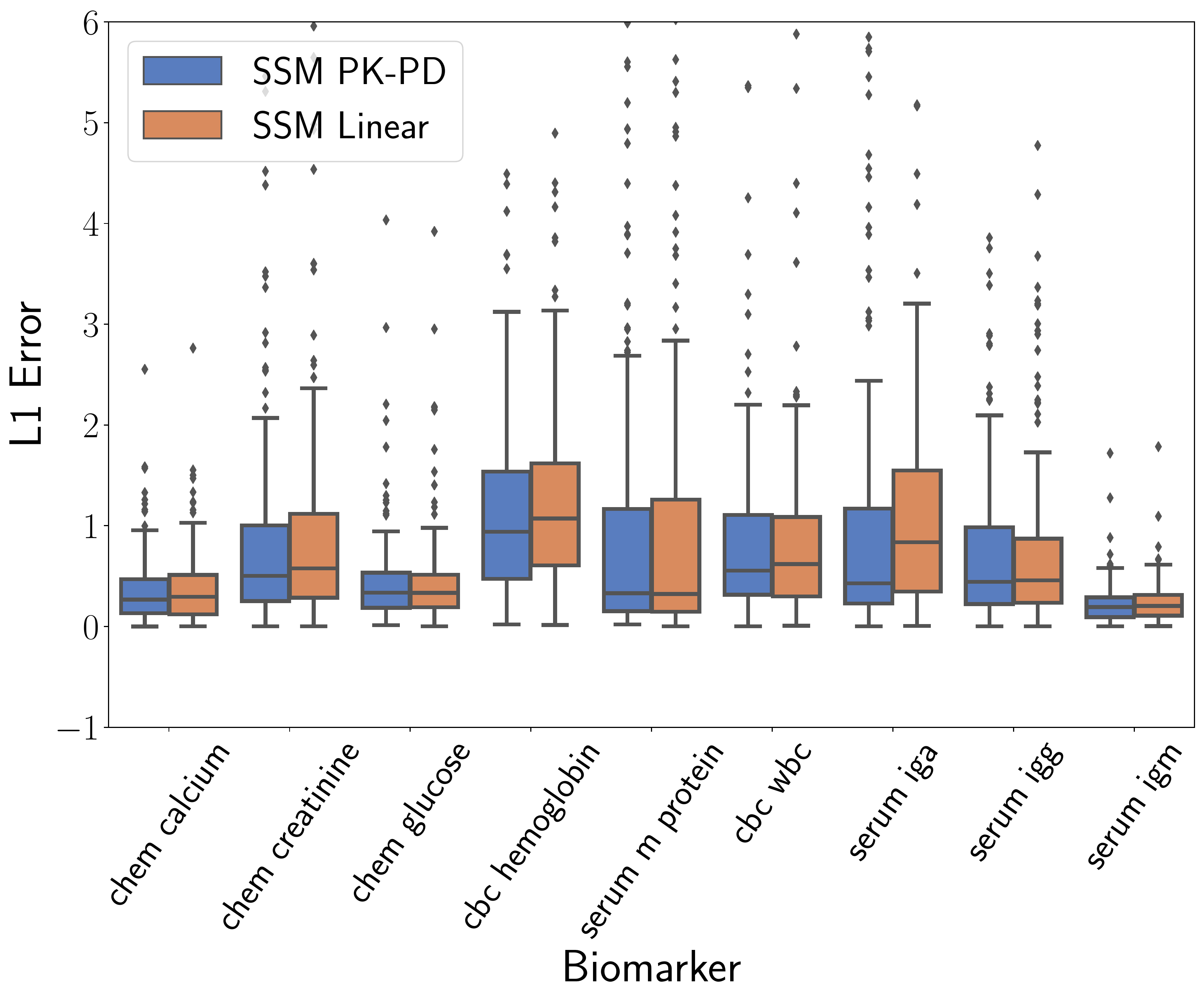}
    \caption{\small \textbf{L1 error for 6-month forward samples from PK-PD and Linear models}: We report L1 error for forward samples over a 6-month time window conditioned only on baseline data.}
    \label{fig:l1_short_subset}
\end{figure}

In Supp. Figure \ref{fig:bar_raw}b), c), and d), we show the L1 error of $\SSMpkpd$ and $\SSMlinear$ when predicting future values of each covariate. We do so under three different conditioning strategies: 1) condition on 6 months of patient data, and predict 1 year into the future; 2) condition on 6 months of patient data, and predict 2 years into the future; 3) condition on 2 years of patient data, and predict 1 year into the future. Observing 1) and 2) (Supp. Figure \ref{fig:bar_raw}b) and c)), we see that prediction quality expectedly degrades when trying to forecast longer into the future. 
Additionally, we find that when increasing the amount of data we condition on to two years (i.e. forward sampling later on in a patient's disease course) (Supp. Figure \ref{fig:bar_raw}d)), the prediction quality is similar to that of conditioning only on six months of data (Supp. Figure \ref{fig:bar_raw}b)) [barring serum M-protein and glucose]. This result reflects the ability of our model to generate accurate samples at multiple stages of a patient's disease. Finally, we also report L1 error for forward samples taken from $\SSMpkpd$ and $\SSMlinear$ over 6-month time windows in Supp. Figure \ref{fig:l1_short_subset}. We see that for some biomarkers, such as serum IgA, the PK-PD model has lower L1 error, while for others, such as Calcium, both models do very well.

\begin{figure}[ht]
\centering
\includegraphics[width=0.48\textwidth]{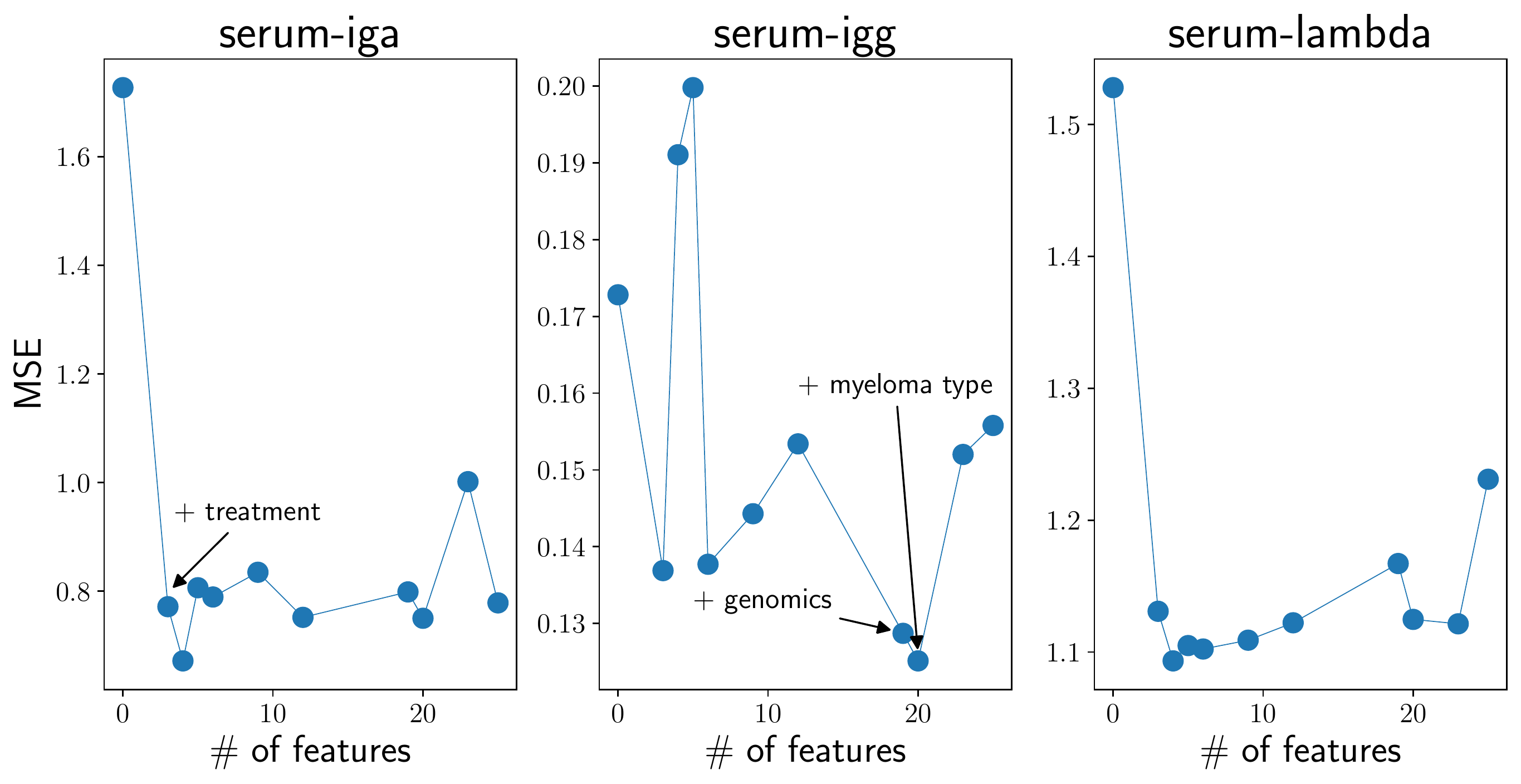}
\caption{\small \textbf{Feature ablation of conditioning set for subset of biomarkers}: We report the per-biomarker mean-squared error (MSE), averaged over all patients and all time points, of $\SSMpkpd$ models trained on an increasing subset of baseline and treatment features.}
\label{fig:abl_mse}
\end{figure}

\subsection{Ablation Studies for $\SSMpkpd$ \label{sec:C5}}
\begin{table*}
    \centering
    \begin{tabular}[t]{c c c c c c c}
    \toprule
     Dataset &\textbf{\begin{tabular}[c]{@{}c@{}}Held-out \\ NELBO \end{tabular}} &    \textbf{ \begin{tabular}[c]{@{}c@{}}Fold 1\end{tabular}} &  \textbf{\begin{tabular}[c]{@{}c@{}}Fold 2\end{tabular}} & \textbf{\begin{tabular}[c]{@{}c@{}}Fold 3\end{tabular}} &
     \textbf{\begin{tabular}[c]{@{}c@{}}Fold 4\end{tabular}} & \textbf{\begin{tabular}[c]{@{}c@{}}Fold 5\end{tabular}} \\
    \midrule                                                    
        \begin{tabular}[c]{@{}c@{}}ML-MMRF\end{tabular} & \text{linear} &       \begin{tabular}[c]{@{}c@{}} 85.26 \end{tabular} & \begin{tabular}[c]{@{}c@{}} 73.25 \end{tabular} &
         \begin{tabular}[c]{@{}c@{}}70.61\end{tabular} & 59.86 & 71.11\\
         & \text{linear + log-cell} &      \begin{tabular}[c]{@{}c@{}}84.42\end{tabular} & \begin{tabular}[c]{@{}c@{}}68.11\end{tabular} &
         \begin{tabular}[c]{@{}c@{}}66.69\end{tabular} & 58.83 & 71.48\\
         & \text{linear + log-cell + te} &     \begin{tabular}[c]{@{}c@{}}65.06\end{tabular} & \begin{tabular}[c]{@{}c@{}}57.20\end{tabular} &
         \begin{tabular}[c]{@{}c@{}}66.73\end{tabular} & 53.37 & 58.12\\
    \bottomrule

    \end{tabular}
    \caption{\small \textbf{Ablation Experiment on ML-MMRF Dataset}: We study the effect of adding each mechanism function to $\SSMpkpd$. We report held-out bounds on negative log likelihood.}
    \label{tab:ablation}
\end{table*}

We report an ablation experiment in Supp. Table \ref{tab:ablation}, where we assess the effect of adding each mechanism function to $\SSMpkpd$ on held-out NELBO. We see that the Neural Log-Cill Kill function gives a modest improvement, while the addition of the Neural Treatment Exponential function gives most of the improvements. 

Secondly, in Figure \ref{fig:abl_mse}, we show a feature ablation experiment to determine the importance of baseline and treatment features in forecasting several multiple myeloma markers. We train $\SSMpkpd$ models on subsets of features, while tuning the latent variable size ($[16,48,64,128]$) on a validation set for each subset. Then, we evaluate the mean-squared error (MSE), averaged over all examples and time points, of each trained model on a separate held-out set. Our results are shown in Figure \ref{fig:abl_mse}. We focus on serums IgA, IgG, and lambda, three biomarkers that are commonly tracked in multiple myeloma to evaluate response to treatment and overall progression of disease \citep{larson2012prevalence,international2003criteria}.

We find that for serums IgA and Lambda, adding the treatment signal intuitively leads to a reduction in the MSE. For serum IgG, while the treatment signal helps with predictive performance, the baseline features, such as the genomic and myeloma type features, also seem to play a role.

\clearpage

\end{document}